\documentclass{article} 
\usepackage{iclr2025_conference,times}

\usepackage{amsmath,amsfonts,bm}

\def\eqref#1{equation~\ref{#1}}

\def\1{\bm{1}}

\DeclareMathAlphabet{\mathsfit}{\encodingdefault}{\sfdefault}{m}{sl}
\SetMathAlphabet{\mathsfit}{bold}{\encodingdefault}{\sfdefault}{bx}{n}

\usepackage{hyperref}
\usepackage{url}
\newcommand{\Sys}{TidalDecode\xspace}
\newcommand{\sa}{PPSA\xspace}
\newcommand{\TODO}{\textcolor{red}{TODO}}
\newcommand{\ZJ}[1]{\textcolor{purple}{ZJ: #1}}

\hypersetup{
  colorlinks,
  citecolor=violet,
  linkcolor=red,
  urlcolor=blue}

\usepackage{microtype}
\usepackage{graphicx}
\usepackage{booktabs} 

\usepackage{multirow,hhline,graphicx,array}
\usepackage{xltabular}
\usepackage{xspace}
\usepackage{float}
\usepackage{algorithm}
\usepackage{algpseudocode}
\usepackage{amsmath}
\usepackage{bbm}
\usepackage{pifont}
\usepackage{titlesec}
\usepackage{adjustbox}
\usepackage{subcaption}
\usepackage{cleveref}
\usepackage{ulem}
\usepackage{lipsum}

\title{TidalDecode: Fast and Accurate LLM Decoding with Position Persistent Sparse Attention}

\author{  
Lijie Yang\thanks{Equal contribution} \\
  Carnegie Mellon University\\
  \texttt{lijiey@andrew.cmu.edu}
    \And
 Zhihao Zhang$^*$ \\
  Carnegie Mellon University\\
  \texttt{zhihaoz3@cs.cmu.edu}
  \And
  Zhuofu Chen \\
  Carnegie Mellon University\\
  \texttt{zhuofuc@cs.cmu.edu} \\
  \And
  Zikun Li \\
  Carnegie Mellon University\\
  \texttt{zikunl@cs.cmu.edu} \\
  \And
  Zhihao Jia \\
  Carnegie Mellon University\\
  \texttt{zhihao@cmu.edu} \\
}

\newcommand{\C}[1]{\Comment{\textcolor{violet}{#1}}}

\iclrfinalcopy 
\begin{document}

\maketitle
\begin{abstract} 
Large language models (LLMs) have driven significant advancements across diverse NLP tasks, with long-context models gaining prominence for handling extended inputs. However, the expanding key-value (KV) cache size required by Transformer architectures intensifies the memory constraints, particularly during the decoding phase, creating a significant bottleneck. 
Existing sparse attention mechanisms designed to address this bottleneck have two limitations: (1) they often fail to reliably identify the most relevant tokens for attention, and (2) they overlook the spatial coherence of token selection across consecutive Transformer layers, which can lead to performance degradation and substantial overhead in token selection. 
This paper introduces \Sys, a simple yet effective algorithm and system for fast and accurate LLM decoding through position persistent sparse attention. \Sys leverages the spatial coherence of tokens selected by existing sparse attention methods and introduces a few token selection layers that perform full attention to identify the tokens with the highest attention scores, while all other layers perform sparse attention with the pre-selected tokens. 
This design enables \Sys to substantially reduce the overhead of token selection for sparse
attention without sacrificing the quality of the generated results.
Evaluation on a diverse set of LLMs and tasks shows that \Sys{} closely matches the generative performance of full attention methods while reducing the LLM decoding latency by up to $2.1\times$\footnote{The codebase to reproduce performance and efficiency results for \Sys included in this paper can be found at \href{https://github.com/DerrickYLJ/TidalDecode}{https://github.com/DerrickYLJ/TidalDecode}}.
\end{abstract}
\section{Introduction}
\label{sec:intro}

Large language models (LLMs) have revolutionized natural language processing (NLP) by achieving state-of-the-art performance on various applications.
As LLMs evolve, they are increasingly being adapted to manage tasks with long contexts, such as Chain-of-Thought reasoning \citep{wei2023chainofthoughtpromptingelicitsreasoning}, document summarization \citep{huang-etal-2021-efficient}, and retrieval-augmented generation \citep{ram2023incontextretrievalaugmentedlanguagemodels, pmlr-v235-zhang24cq}.
However, quickly and efficiently serving long-context LLMs is challenging due to the inherent memory and compute bottlenecks in the Transformer architectures~\citep{vaswani2023attentionneed}.

LLM inference involves two separate stages: {\em prefilling} and {\em decoding}. 
The prefilling stage computes the activations for all input tokens and stores the keys and values for all tokens in the {\em key-value (KV) cache}, allowing the LLM to reuse these keys and values to compute attention for future tokens.
In each decoding stage, the LLM decodes one new token using all input tokens and previously generated tokens.
The KV cache size grows linearly in the sequence length~\citep{kwon2023efficient}.
For instance, with a context length of 128K tokens, the KV cache of LLama2-7B with half-precision can easily reach 64 GB\footnote{The KV cache size is computed as: 
$\text{Layers} \times \text{KV Heads} \times \text{Head Dim} \times \text{Seq Len} \times \text{FP16 Size} \times 2 \ (\text{for K+V}) = 32 \times 32 \times 128 \times 128K \times 2 \ \text{bytes} \times 2 = 64 \ \text{GB}$.}, creating substantial memory pressure for LLM serving.
In addition, the LLM decoding stage is memory-bounded since decoding one new token requires accessing all previous tokens in the KV cache, making KV cache access the primary bottleneck for long-context LLM decoding. 
This memory-bound nature severely limits the scalability and efficiency of LLM serving.

To address this problem, recent work has introduced {\em sparse attention}, which approximates full attention using a small portion of tokens with the highest attention scores. Compared to full attention, sparse attention reduces computation cost and memory access while preserving the LLM's generative performance~\citep{ge2024model, zhang2023h2o}.
Existing sparse attention techniques can be classified into two categories: eviction- and selection-based methods.

First, {\em eviction-based} sparse attention reduces memory usage for the KV cache by selectively discarding less relevant tokens from the KV cache, therefore reducing the number of tokens computed in attention mechanisms~\citep{xiao2023streamingllm, zhang2023h2o}. 
While these methods decrease the size of the KV cache, they can be inadequate for tasks where critical information is carried by tokens that are prematurely evicted, such as the needle-in-the-haystack tasks~\citep{peng2023yarn}.
On the other hand, {\em selection-based} sparse attention maintains {\em all} tokens in the KV cache, estimates their attention scores, and selects a small subset of tokens to participate in each LLM decoding step.
This approach is prone to issues related to distribution shifts caused by appending sparsely attended, biased KV representations back into the cache.

\begin{figure}
    \centering
    \includegraphics[width=0.75\textwidth]{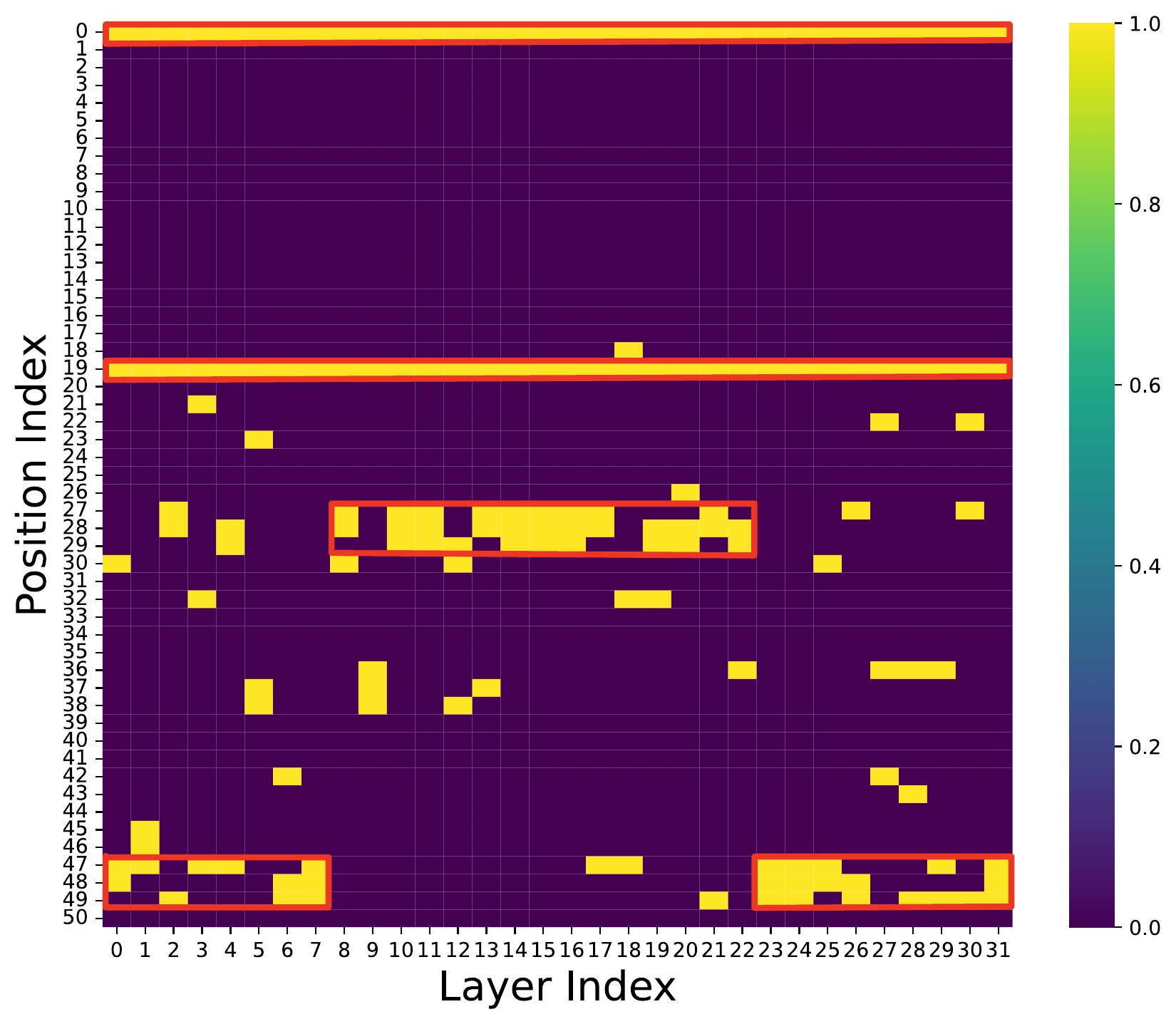}
    \caption{The heatmap for one decoding step of Llama3-8B-Instruct~\citep{gradient-ai-llama-3-8B}, where columns and rows indicate different Transformer layers and tokens in the KV cache, respectively. For each layer, the 5 tokens (10\% sparsity) with the highest attention scores of the first attention head are highlighted in yellow, which are the tokens used for sparse attention.
    We feed an input prompt ``Use only the provided search results to write a high-quality, concise answer to the question.\textbackslash n\textless|begin\_of\_text|\textgreater\textbackslash n The magic number is: 15213. \textbackslash n\textbackslash n\textbackslash n Question: What is the magic number? Keep the response short and direct. Answer: '', and the LLM outputs ``15213''. The results show strong spatial coherence of tokens chosen for sparse attention in the decoding step.}
    \label{fig:PPSA_heatmap}
\end{figure}

This paper presents \Sys, an algorithm and system for fast and precise LLM decoding, utilizing {\em position persistent sparse attention} (\sa). 
A key insight behind \Sys is the observation that tokens chosen for sparse attention --- based on their highest attention scores --- exhibit {\em significant overlap across consecutive Transformer layers} within each decoding phase.
\Cref{fig:PPSA_heatmap} illustrates this overlap in a single decoding step of LLaMA-3-8B instruct~\cite{gradient-ai-llama-3-8B} with an input of 51 tokens. 
Each column in the figure corresponds to a Transformer layer, and each row indicates one token in the KV cache.
Selection-based sparse attention methods select the 5 tokens with the highest attention scores (highlighted in yellow) for attention computation in each head.
As the figure depicts, there is a recurring pattern where consecutive layers {\em consistently focus on the same set of tokens}, indicating a {\em spatial coherence in the selection of tokens} for sparse attention.

Instead of independently selecting tokens for sparse attention at each layer, \Sys introduces a few token selection layers, which perform full attention to identify the tokens with the highest attention scores. 
All remaining layers implement position persistent sparse attention, where only the tokens selected by the token selection layers are retrieved from the KV cache for attention.
Consequently, all other layers between two token selection layers operate on the same set of tokens, reducing the overhead of token selection. 
Experiments across a diverse set of LLMs and datasets demonstrate that using just two token selection layers --- one at the beginning and one in the middle --- is sufficient to achieve high generative performance while minimizing computation and memory overheads.

This design enables \Sys to substantially reduce the overhead of token selection for sparse attention without sacrificing the quality of the generated results. Additionally, to address the KV cache distribution shift, \Sys introduces a cache-correction mechanism that periodically refills the KV cache using full attention for all sparsely decoded tokens to mitigate bias in the KV representations.

Comprehensive evaluation with the LongChat-7b-v1.5-32k, Llama-3-8B, Llama-3-70B, and Llama-3.1-8B models on the Needle-in-the-Haystack, PG-19, and LongBench tasks demonstrates that \Sys{} can consistently achieve the best performance efficiency trade-off compared with the best existing sparse attention methods.
We have implemented custom GPU kernels for \sa and an end-to-end system for \Sys{}. Compared with existing full and sparse attention implementations, our system reduced the end-to-end inference latency by up to $2.1\times$ and $1.2\times$, respectively.
In conclusion, our contributions are:
\begin{itemize}
    \item We propose \Sys{}, a streamlined and efficient algorithm and system for fast and high-quality LLM decoding, utilizing position persistent sparse attention. 
    \item To address KV cache distribution shifts, we introduce a cache-correction mechanism that periodically refills the KV cache with using full attention for sparsely decoded tokens. 
    \item Empirically, we demonstrate the effectiveness and efficiency of \Sys{} through comprehensive evaluation, showing that \Sys significantly outperforms existing sparse attention methods.
\end{itemize}
\section{Related Work}
\label{sec:related_work}
\paragraph{Long-context model.}
Efficiently handling long-context inputs is essential for various LLM tasks in real-world applications such as document summarization, question answering, and dialogue systems~\citep{wang2024limitssurveytechniquesextend}.
Recent advancements, including rotary positional encoding (RoPE) \citep{su2023roformer}, have enabled models to manage extended context lengths effectively. 
The LLaMA-3 model series supports up to 8K tokens, with enhanced versions such as Gradient-AI-Llama3 \citep{gradient-ai-llama-3-8B} and LLaMA 3.1 \citep{llama3-1} extending this limit to 128K tokens. 
Additionally, proprietary LLMs such as GPT-4 Turbo and GPT-4o \citep{gpt4} support up to 128K tokens, and Claude 3.5 Sonnet allows up to 200K tokens \citep{claude}.
While recent work has introduced efficient attention kernel implementation \citep{dao2022flashattention, dao2023flashattention},  processing long-context inputs continues to be constrained by significant memory usage and computational costs from the extended KV cache.
\Sys{} is designed to mitigate these challenges by reducing latency and memory overhead through an efficient strategy for selecting tokens with the highest attention scores and one-time intermediate re-calibration, ensuring both efficiency and high-quality output.

To alleviate the intrinsic computational and memory bottleneck in long-context LLM inference, recent works on sparse attention have approached this problem from two main perspectives: {\em eviction-} and {\em selection-based} methods.

\paragraph{Eviction-based sparse attention.} \citet{xiao2023streamingllm, zhang2024h2o} propose to reduce KV cache memory usage by evicting tokens that are considered less relevant during inference. 
These suffer from potential performance degradation, especially in tasks where every token may carry crucial information (e.g., needle-in-the-haystack tasks), since tokens with high importance for a future decoding step can be mistakenly evicted as the generation proceeds, which makes {\em selection-based methods} more popular choices in latest sparse attention works.

\paragraph{Selection-based sparse attention.} Instead of evicting past tokens in the KV cache, \citet{child2019generating, kitaev2020reformer, choromanski2020rethinking, tang2024questqueryawaresparsityefficient, ribar2023sparq} preserve the full KV cache and only select important tokens to attend with the attention module on the fly. More specifically, \citet{child2019generating} leverages a fixed attention mask to select tokens while \citet{tang2024questqueryawaresparsityefficient, ribar2023sparq, choromanski2020rethinking, kitaev2020reformer} aim to identify and retain the most relevant tokens at each layer by approximating attention scores.
Although these methods are more selective, they operate independently at each layer and are not guaranteed to obtain the ground-truth tokens with the highest attention scores, failing to capture token relevance patterns that persist across layers. 
Moreover, attention score estimation algorithms sometimes introduce unnecessary complexity, diminishing the practical efficiency gains they are designed to achieve.
Improving upon prior works, \Sys{} leverages a shared pattern of most important tokens across consecutive layers to further reduce the computational overhead and memory access required for token selection. 
\section{Methodology}
\label{sec:methodology}
\begin{figure}
    \centering
    \includegraphics[width=0.95\textwidth]{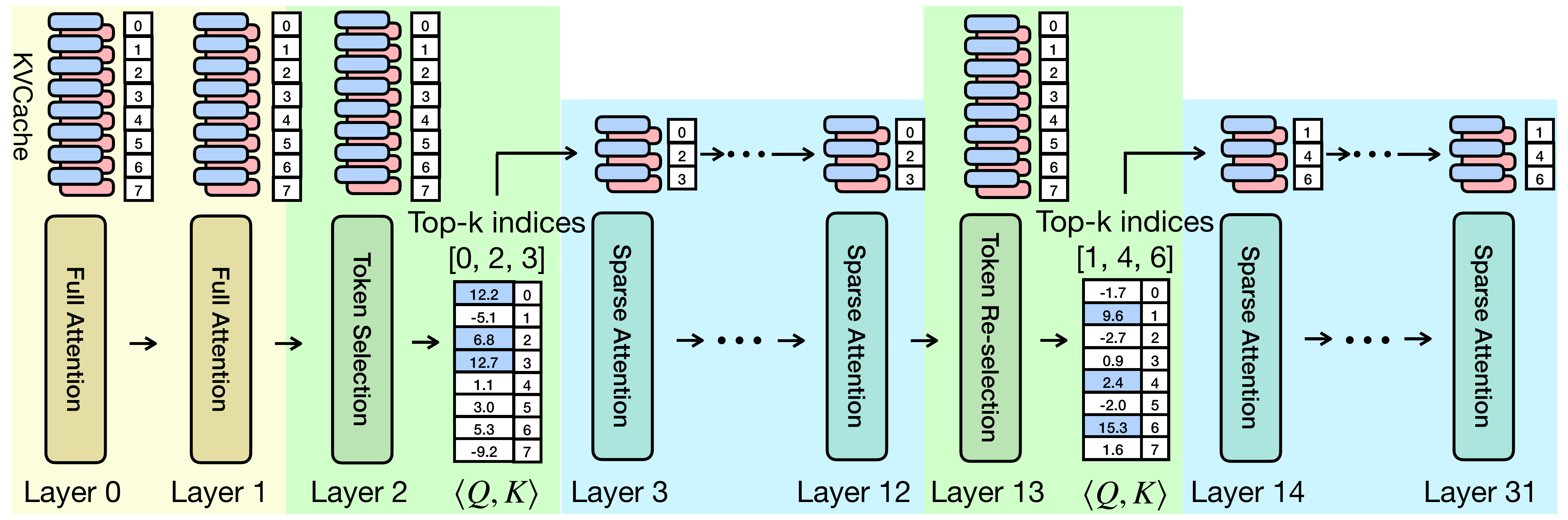}
    \caption{An overview of the decoding step in \Sys, which performs full attention for the first two layers, full attention with token selection for the third layer and a middle layer, and position persistent sparse attention for all other layers.}
    \label{fig:arch}
\end{figure}

This section introduces \Sys{}, an efficient algorithm and system for fast LLM decoding using {\em position persistent sparse attention} and {\em KV cache correction}. \Cref{fig:arch} shows an overview of \Sys. 
\Sys uses the same prefilling mechanism as existing systems and performs full attention to compute the key-value (KV) cache for all prompt tokens.
In each decoding step, \Sys uses three types of attention layers: full attention, full attention with token selection, and position persistent sparse attention.
First, \Sys performs full attention for the initial Transformer layers to avoid early performance degradation as identified by prior work~\citep{tang2024questqueryawaresparsityefficient}.
Second, the layer immediately after full attention and a single middle layer (e.g., layer 2 and 13 in \Cref{fig:arch}) perform full attention with token selection, where \Sys stores the inner product\footnote{We don't store attention score as state-of-the-art attention kernels don't materialize the attention score. Since the softmax operation is ordering invariant, we store the inner product value instead.} between the current query and key vectors of all tokens in KV cache during full attention and then selects $k$ tokens contributing to the highest attention scores.
Third, all other layers perform {\em position persistent sparse attention}, where only tokens selected from the previous token selection layer are loaded from the KV cache to perform attention computation.

\subsection{Position Persistent Sparse Attention (\sa)}
Attention mechanisms have been widely used in today's LLMs. For each attention head, the output is computed via scaled multiplicative formulation as follows.
\begin{equation}
A_i = Q_i K_i / \sqrt{d}, \quad H_i = {\rm softmax}(A_i) V_i
\end{equation}
where $Q_i$, $K_i$, and $V_i$ are the query, key, and value tensors for the $i$-th attention head. $A_i$ is a matrix representing the attention scores between tokens, and $H_i$ is the output of the $i$-th attention head. 
Instead of attending to all input tokens, existing sparse attention methods approximate attention computation by attending the query $Q_i$ to a subset of previous tokens the highest attention scores.
Prior work generally performs token selection for individual attention heads and Transformer layers, introducing high runtime overhead. 
For example, selecting the tokens with highest attention scores using top-k can take longer than computing full attention (see~\Cref{fig:kernel23}), thus diminishing the benefits of performing sparse attention.

\begin{figure}[ht]
    \centering
    \begin{subfigure}[b]{0.45\textwidth}
        \includegraphics[width=1\textwidth]{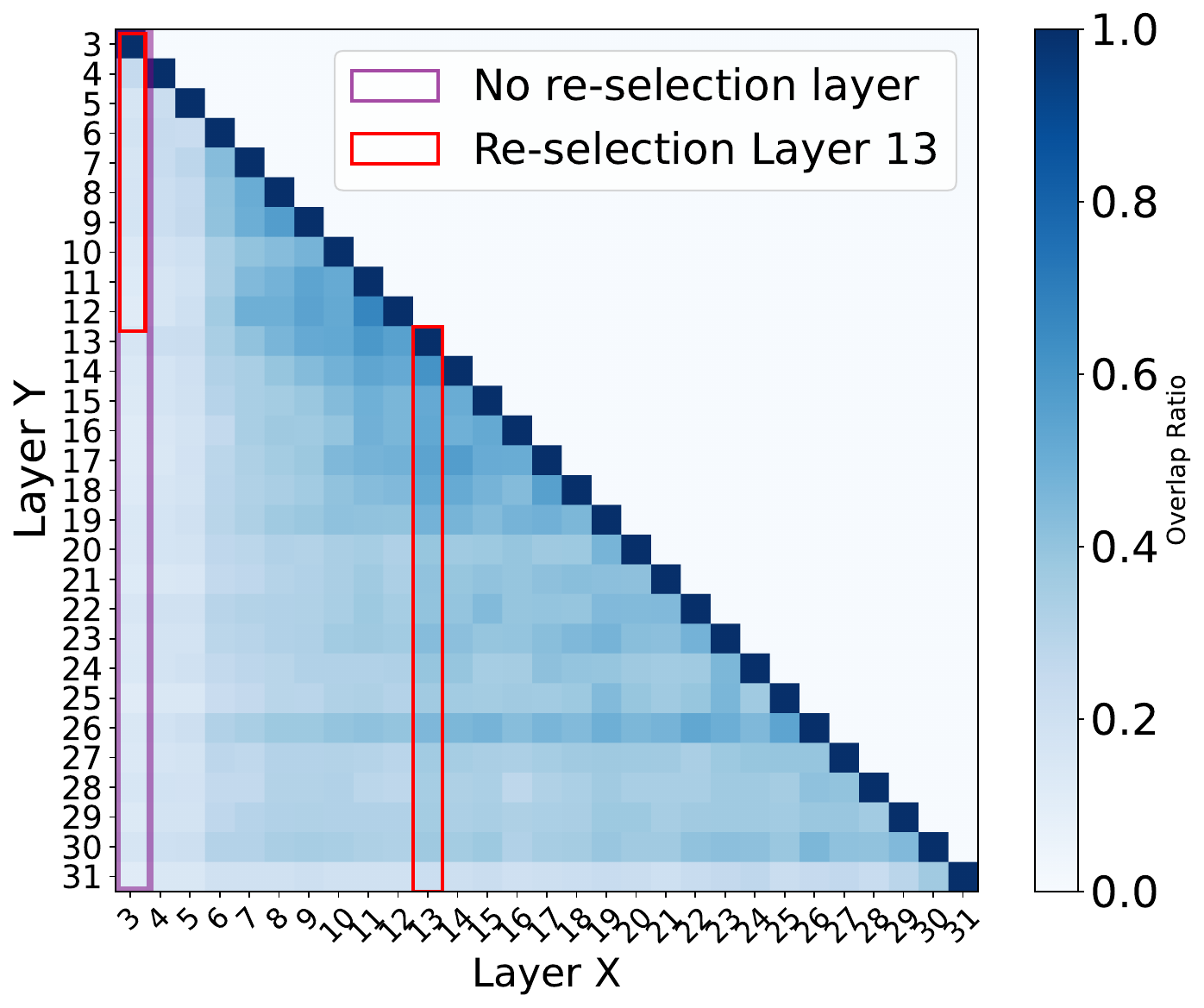}
        \caption{Overlap of Tokens with the Highest Attention Scores between Layers}
        \label{fig:correction_overlap_matrix_256}
    \end{subfigure}
    \hspace{0.05\textwidth} 
    \begin{subfigure}[b]{0.45\textwidth}
        \includegraphics[width=1\textwidth]{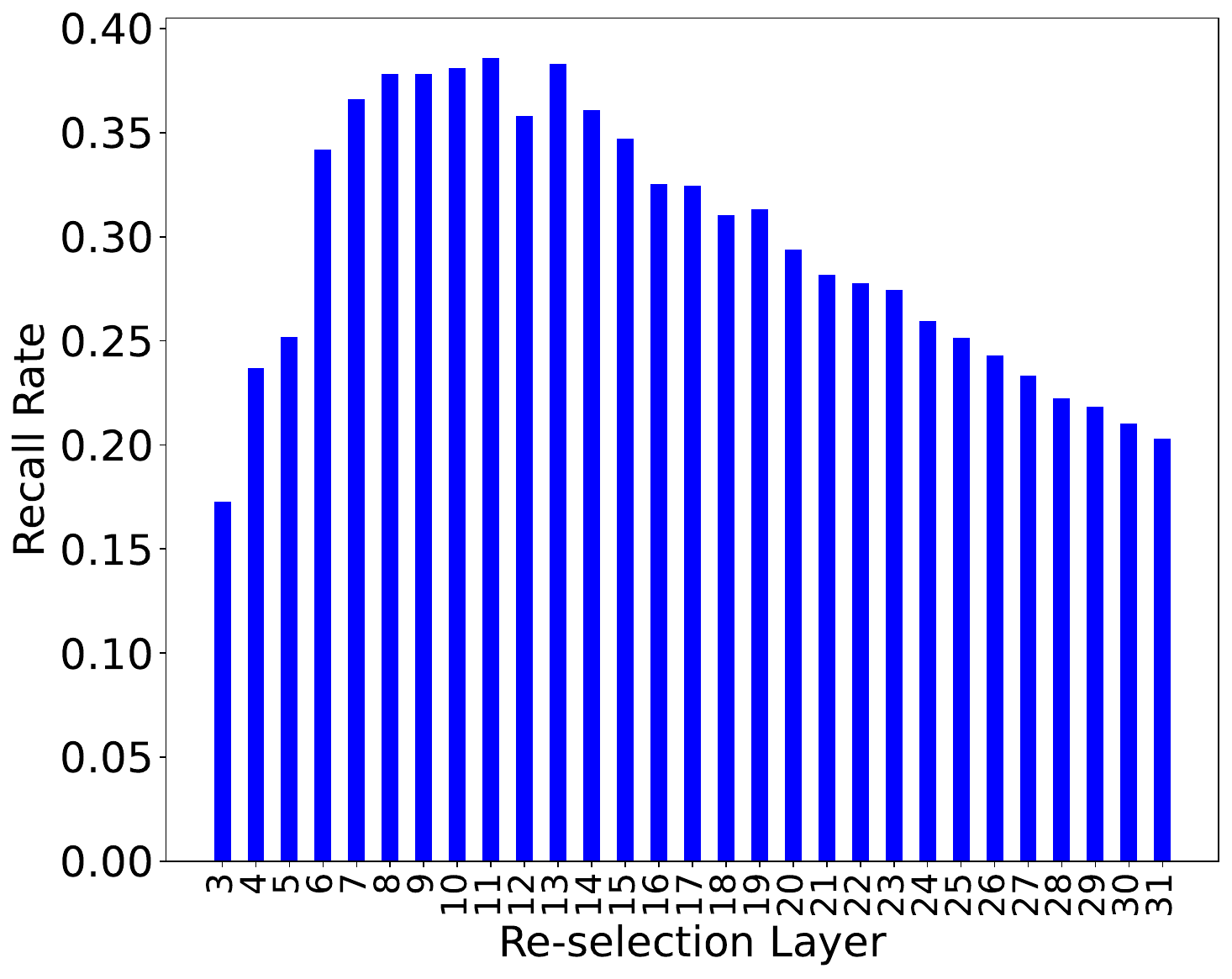}
        \caption{Recall Rate by Re-selection Layer}
        \label{fig:recall_rates_bar_256}
    \end{subfigure}
    \caption{By retrieving the top-256 tokens from a 100K-context-length Needle-in-the-Haystack test conducted on PG-19-mini, \ref{fig:correction_overlap_matrix_256} shows the overlap ratio of tokens with the highest attention scores across layers, showing that consecutive layers tend to share a large number of critical tokens. \ref{fig:recall_rates_bar_256} depicts the recall rates, indicating that different choices of re-selection layers have a high impact on the recall rates --- there is a clear peak, delineating the optimal layers for token re-selection.}
    \label{fig:matrix_recall}
\end{figure}

The key insight behind \Sys's position persistent sparse attention is an observation that {\em tokens with highest attention scores for consecutive Transformer layers highly overlap}.
We use the LLaMA-3-8B model and the needle-in-the-haystack test on PG-19-mini dataset with a context length of 100K tokens to quantify this observation.
We randomly select 100 requests from the dataset, compute full attention, and analyze the top 256 tokens with the highest attention scores for each Transformer layer.
\Cref{fig:correction_overlap_matrix_256} shows the overlap ratios for all pairs of transformer layers, where an overlap ratio of 1 indicates that the tokens with highest attention scores are always identical in these layers, and an overlap ratio of 0 means the top tokens do not overlap in the two layers. Note that we select top 256 tokens from 100K tokens in the KV cache, so randomly selected tokens hardly overlap.
In \Cref{fig:recall_rates_bar_256}, we compute average recall rates of selected tokens by choosing different re-selection layers. We observe that without re-selection layers, where all layers possess a low overlap ratio with Layer 3 shown in the purple box in \Cref{fig:correction_overlap_matrix_256}, the average recall rates are less than 20\%. When we choose Layer 13 to perform re-selection, the average recall rates boost to almost 40\% due to higher overlap ratios between Layer 13 and its subsequent layers, shown by red boxes in \Cref{fig:correction_overlap_matrix_256}. 

\begin{algorithm}
\caption{\Sys{}}
\label{alg:decoding}
\begin{algorithmic}[1]
\State {\bf Input:} Current embedding $h$, KV cache $\mathcal{C}$, token budget $m$
\State {\bf Output:} Logits
\State {\bf Initialize:} $\rho = []$ \C{Initialize the token buffer to store selected tokens}
\For{each decoder layer $i$}
    \State $q, k, v = f(W_{qkv}, h)$
    \State $\mathcal{C}$.append($k, v$)
    \If{$i$ is Full Attention Layer}
        \State $o = \text{FullAttention}(q, \mathcal{C}[:])$ \C{Dense attention with the full KVCache}
    \ElsIf{$i$ is Token Selection Layer}
        \State $o = \text{FullAttention}(q, \mathcal{C}[:])$ \C{Dense attention with the full KVCache}
        \State $K \leftarrow \mathcal{C}.\text{getKey}$, $\rho:=\text{argTopK}(\langle q, K\rangle, m)$ \C{Update token buffer}
    \Else
        \State $o = \text{SparseAttention}(q, \mathcal{C}[\rho])$ \C{Sparse attention with the tokens in the token buffer}
    \EndIf
    \State $h = \text{FFN}(o)$
\EndFor
\State logits $=$ lm\_head$(h)$
\State \Return logits
\end{algorithmic}
\end{algorithm}

Based on this observation, we design position persistent sparse attention to maximally leverage the token overlaps between consecutive Transformer layers to reduce the computation cost for token selection while achieving high predictive performance. \Cref{alg:decoding} shows the \Sys algorithm for interleaving full attention and \sa layers. 
After the initial full attention layers, \Sys uses a token selection layer that computes full attention and selects tokens with the highest attention scores. 
To select tokens, \Sys{} stores the inner product $\langle Q, K \rangle$ on the fly together with full attention calculation. \Sys then selects the top $k$ tokens with the highest inner product values to form a token set $\mathcal{T}$. Note that using the inner product to select top-$k$ is equivalent to the post-softmax attention score as the softmax operator is ordering invariant.  
All \sa layers after a token selection layer computes sparse attention by only loading the keys and values for tokens in $\mathcal{T}$, thus limiting the number of tokens participating in attention computations and reducing memory access.

A straightforward approach to designing \Sys is to select the tokens $\mathcal{T}$ once after full attention and perform \sa using the same set of tokens for all subsequent layers.
However, our preliminary experimentation shows that using a single token set for all Transformer layers reduces the LLM's predictive performance by a large margin since distant Transformer layers are less correlated compared to consecutive layers, as shown in \Cref{fig:correction_overlap_matrix_256}.
To address this issue, \Sys performs {\em token re-selection} in a middle layer, where \Sys{} recalibrates the selected tokens with the highest attention scores by applying full attention and re-selecting top-$k$ token to update $\mathcal{T}$, ensuring that token selection remains optimal for the remaining layers. 
This re-selection mechanism significantly boosts the model performance and promotes accurate and efficient \sa throughout the model.

Extensive evaluation on both small and large models on a wide range of datasets shows that using a single middle layer for token re-selection is sufficient to preserve the LLM's generative performance, while introducing small runtime overhead.
However, deciding which layer to perform token re-selection is critical to model performance.
As shown in \Cref{fig:matrix_recall}, choosing different layers for token re-selection results in different recall rates, where layer 11 and 13 achieve optimal performance.
Introducing a one-time token re-selection at an optimal layer ensures the selected tokens are re-calibrated, effectively mitigating the drift in token importance and elevates accuracy from 15\% (without re-selection) to almost 40\%.

\subsection{KV Cache Correction}
\begin{figure}
    \centering
    \includegraphics[width=0.6\textwidth]{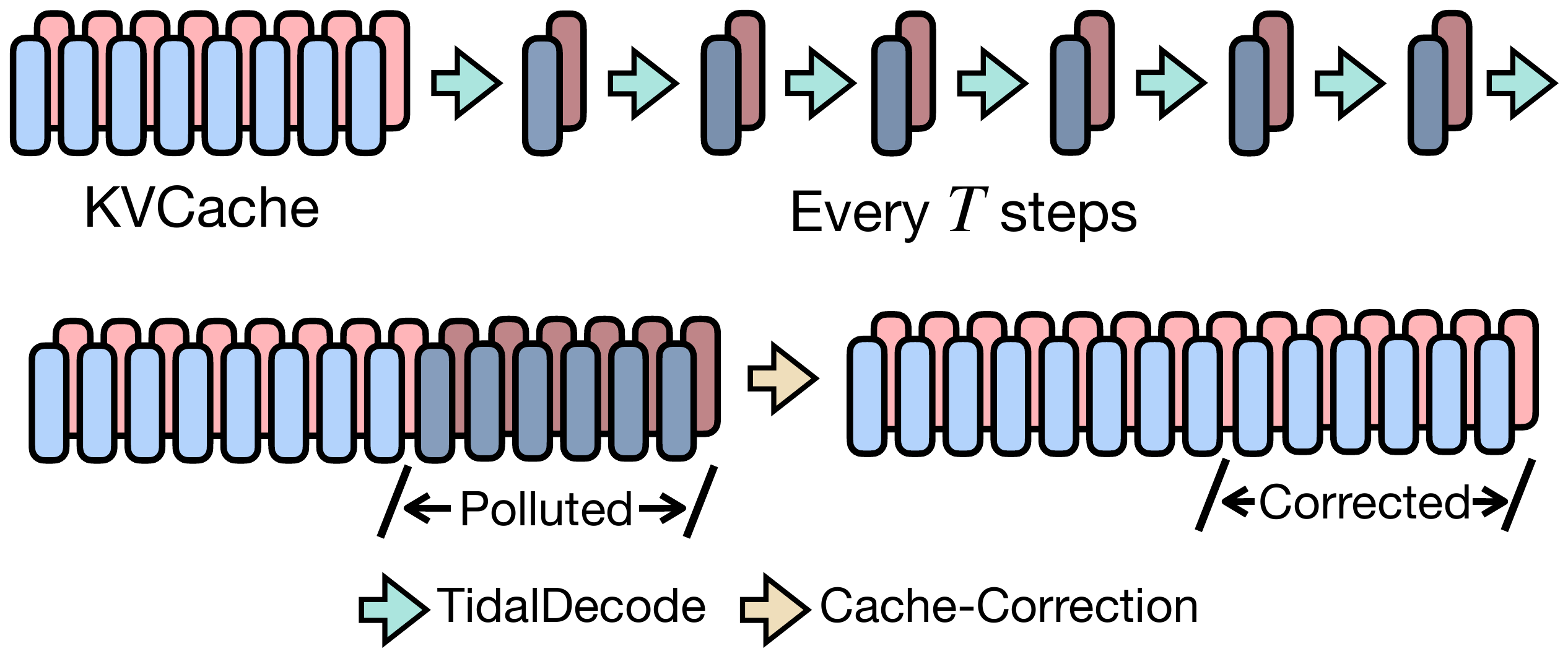}
    \caption{Cache Correction}
    \label{fig:cache-correction}
\end{figure}

For tokens decoded by sparse attention methods, their key/value representations can deviate from the original representation of full attention decoded ones, which we refer to as polluted tokens. The problem can be further exacerbated as their KV pairs are added to the KV cache, resulting in the error accumulation or distribution shift of the KV cache. This can lead to model performance drop in scenarios where the generation length is fairly long. To this end, \Sys{} uses a cache-correction mechanism as shown in~\Cref{fig:cache-correction} to periodically correct the polluted tokens in the KV cache. For every $T$ decoding step performed by \Sys{}, there will be a cache correction step through a prefill over all polluted tokens to update their KV representations in the cache. The choice of $T$ can be at the level of thousands of decoding steps but also depend on different models and tasks. Notice that the cache correction step can be performed concurrently with the sparse decoding step. Nevertheless, we haven't used cache correction in our evaluations to make it a fair comparison against existing methods.

\section{Experiments}
\label{sec:experiments}

\subsection{Experiment Setting}
In this section, we conduct extensive experiments to assess both the performance and efficiency of \Sys{}. 
Our evaluations are performed on widely used open-source models, including Llama-2-7B \cite{touvron2023llama} and Llama-3-8/70B. 
Both models are pretrained decoder-only transformers, exhibiting similar yet distinct architectural features. 
For instance, Llama 3-8B incorporates group query attention (GQA), a feature not present in Llama 2-7B. In~\Cref{sec:performance-eval}, we evaluate \Sys{}'s performance on various tasks, including needle-in-the-haystack, language modeling on PG-19, and LongBench. In~\Cref{sec:efficiency-eval}, we write customized attention kernels and compare \Sys{}'s kernel efficiency against existing state-of-the-art sparse attention methods. Finally, in~\Cref{sec:ablation-eval}, we conclude our evaluations with a detailed sensitivity analysis on the choice of different token selection layers. We use TD+LX to denote \Sys{} with layer X selected as the token re-selection layer throughout this section.
\subsection{Performance Evaluation}
\label{sec:performance-eval}
To evaluate the effectiveness of \Sys{}, we conduct two key downstream NLP experiments: the needle-in-the-haystack test and perplexity evaluation on the PG-19 dataset~\citep{raecompressive2019}. These tasks provide robust benchmarks for measuring both sparse attention models' ability to retrieve critical information in challenging scenarios and their performance on long-context language modeling tasks.
\subsubsection{Needle-in-the-Haystack}

\begin{table}[ht]
\centering
\caption{Results of 10k-context-length Needle-in-the-Haystack test on LongChat-7b-v1.5-32k. \Sys{} achieves the same or better results than Quest and significantly better results than cache eviction algorithms such as H2O, TOVA, and StreamingLLM. \Sys{} achieves full accuracy with only a 512 token budget.}
\begin{tabular}{crrrrr}
    \toprule
    Method / Budget & K=32 & K=64 & K=128 & K=256 & K=512 \\
    \midrule
    H2O & 0\% & 1\% & 1\% & 1\% & 3\% \\
    TOVA & 0\% & 1\% & 1\% & 3\% & 8\% \\
    StreamingLLM & 1\% & 1\% & 1\% & 3\% & 5\% \\
    Quest & 65\% & \textbf{99\%} & \textbf{99\%} &\textbf{99\%} & \textbf{100\%} \\
    TD+L7(Ours) & \textbf{73\%} & 92\% & 98\% & \textbf{99\%} & \textbf{100\%} \\
    \bottomrule
    \end{tabular}
    \label{tab:longchat-7b-results}
\end{table}

\begin{table}[]
\centering
\caption{Comprehensive results of 10K-, 32K-, and 100K-context-length Needle-in-the-Haystack test on Llama-3-8B-Instruct-Gradient-1048k, Llama-3.1-8B-Instruct, and Llama-3-70B-Instruct-Gradient-1048k with PG-19-mini dataset. Across all models, \Sys{} consistently outperforms Quest, showing that \Sys{} with only two token selection layers can effectively retain critical information. \Sys{} achieves full accuracy with 64, 64, and 128 tokens in 10K-, 32K-, and 100K-context-length tests, which is only 0.6\%, 0.2\%, and 0.1\% of total input lengths, respectively.}
\begin{tabular}{clrrrrr}
\toprule
Model (context length)                                                                       & Method / Budget & K=32           & K=64           & K=128          & K=256                   & K=512          \\ \midrule
\multirow{3}{*}{\begin{tabular}[c]{@{}c@{}}LLaMA-3-8B\\ (10K)\end{tabular}}   & Quest         & 74\%           & 84\%           & 99\%           & 98\%                    & \textbf{100\%} \\
& TD+L13(Ours)        & 88\%           & \textbf{98\%}  & \textbf{100\%} & \textbf{100\%}          & \textbf{100\%} \\
& TD+L15(Ours)        & \textbf{92\%}  & 88\%           & 94\%           & 94\%                    & \textbf{100\%} \\ \midrule
\multirow{3}{*}{\begin{tabular}[c]{@{}c@{}}LLaMA-3-8B\\ (100K)\end{tabular}}  & Quest         & 38\%              & 50\%           & 65\%           & 87\%                    & 98\%           \\
& TD+L13(Ours)        & \textbf{86\%}              & \textbf{92\%}  & \textbf{100\%} & \textbf{100\%}          & \textbf{100\%} \\
& TD+L15(Ours)        & 84\%             & 90\%           & 92\%           & 98\%                      & \textbf{100\%} \\ \midrule
\multirow{3}{*}{\begin{tabular}[c]{@{}c@{}}LLaMA-3.1-8B\\ (10K)\end{tabular}} & Quest         & 74\%           & 86\%           & 94\%           & \textbf{100\%}          & 98\%           \\
& TD+L13(Ours)        & \textbf{100\%} & \textbf{100\%} & \textbf{100\%} & \textbf{100\%}          & \textbf{100\%} \\
& TD+L14(Ours)        & 98\%           & \textbf{100\%} & \textbf{100\%} & \textbf{100\%}          & \textbf{100\%} \\ \midrule
\multirow{3}{*}{\begin{tabular}[c]{@{}c@{}}LLaMA-3.1-8B\\ (32K)\end{tabular}} & Quest         & 78\%              & 88\%           & 92\%           & \textbf{100\%}          & \textbf{100\%} \\
& TD+L13(Ours)        & \textbf{98\%}              & \textbf{100\%} & \textbf{100\%} & \textbf{100\%}          & \textbf{100\%} \\
& TD+L14(Ours)        & 80\%              & 98\%           & \textbf{100\%} & \textbf{100\%}          & \textbf{100\%} \\ \midrule
\multirow{3}{*}{\begin{tabular}[c]{@{}c@{}}LLaMA-3-70B\\ (10K)\end{tabular}}  & Quest         & 68\%              & 72\%           & 90\%           & 98\%                    & \textbf{100\%} \\
& TD+L14(Ours)        & 87\%           & 93\%           & \textbf{100\%} & \textbf{100\%} & \textbf{100\%} \\
& TD+L31(Ours)        & \textbf{90\%}  & \textbf{97\%}  & \textbf{100\%} & \textbf{100\%} & \textbf{100\%} \\ \midrule
\multirow{3}{*}{\begin{tabular}[c]{@{}c@{}}LLaMA-3-70B\\ (32K)\end{tabular}}  & Quest         & 50\%              & 80\%           & 88\%           & 92\%                    & 78\%           \\
& TD+L14(Ours)        & \textbf{82\%}              & \textbf{98\%}  & \textbf{98\%}  & \textbf{100\%}          & \textbf{100\%} \\
& TD+L31(Ours)        & 80\%            & 82\%           & 92\%           & 98\%                    & \textbf{100\%} \\ \bottomrule
\end{tabular}
\label{tab:needle}
\end{table}
The Needle-in-the-Haystack test assesses LLMs' ability to handle long-dependency tasks, which is particularly critical for sparse attention algorithms. 
Eviction-based methods \cite{xiao2023streamingllm, zhang2023h2o} may discard essential tokens, while selection-based approaches often fail to consistently identify the ground-truth tokens with the highest attention scores in long contexts.
Since Quest is the current state-of-the-art approach on this task, we first run \Sys{} on the same test as Quest on the LongChat-7b-v1.5-32k model and obtained~\Cref{tab:longchat-7b-results} with competitive performance. 
To demonstrate the effectiveness of \Sys{} on long-dependency tasks, we further evaluate \Sys{} on tasks with 10K-, 32K-, and 100K-context-window lengths with the LLaMA-3-70B, LLaMA-3-8B, LLaMA-3.1-8B model using the PG-19-mini dataset, shown in~\Cref{tab:needle}.
To ensure fairness, both \Sys{} and Quest use dense attention in the first two layers. In each test, we inserted a random password within the text and tested whether the specific method could retrieve the password correctly. 

From~\Cref{tab:needle}, \Sys{} consistently outperforms Quest and achieves full accuracy with an extremely low sparsity (about 0.5\% across all context lengths and models). These results demonstrate \Sys{} can achieve state-of-the-art performance with only two token selection layers. 
While Quest relies on page-level importance estimation for token selection, \Sys{}'s exact selection with token reuse approach proves more effective for this task. Also, note that \Sys{} can reduce the token budget by up to $8\times$ when achieving a 100\% accuracy compared with Quest. This further demonstrates that \Sys{}'s exact token selection layer can obtain more relevant tokens than Quest.

\subsubsection{Language Modeling}
\begin{figure}[htp!]
    \centering
    \begin{subfigure}[b]{0.49\textwidth}
        \centering
        \includegraphics[width=\textwidth]{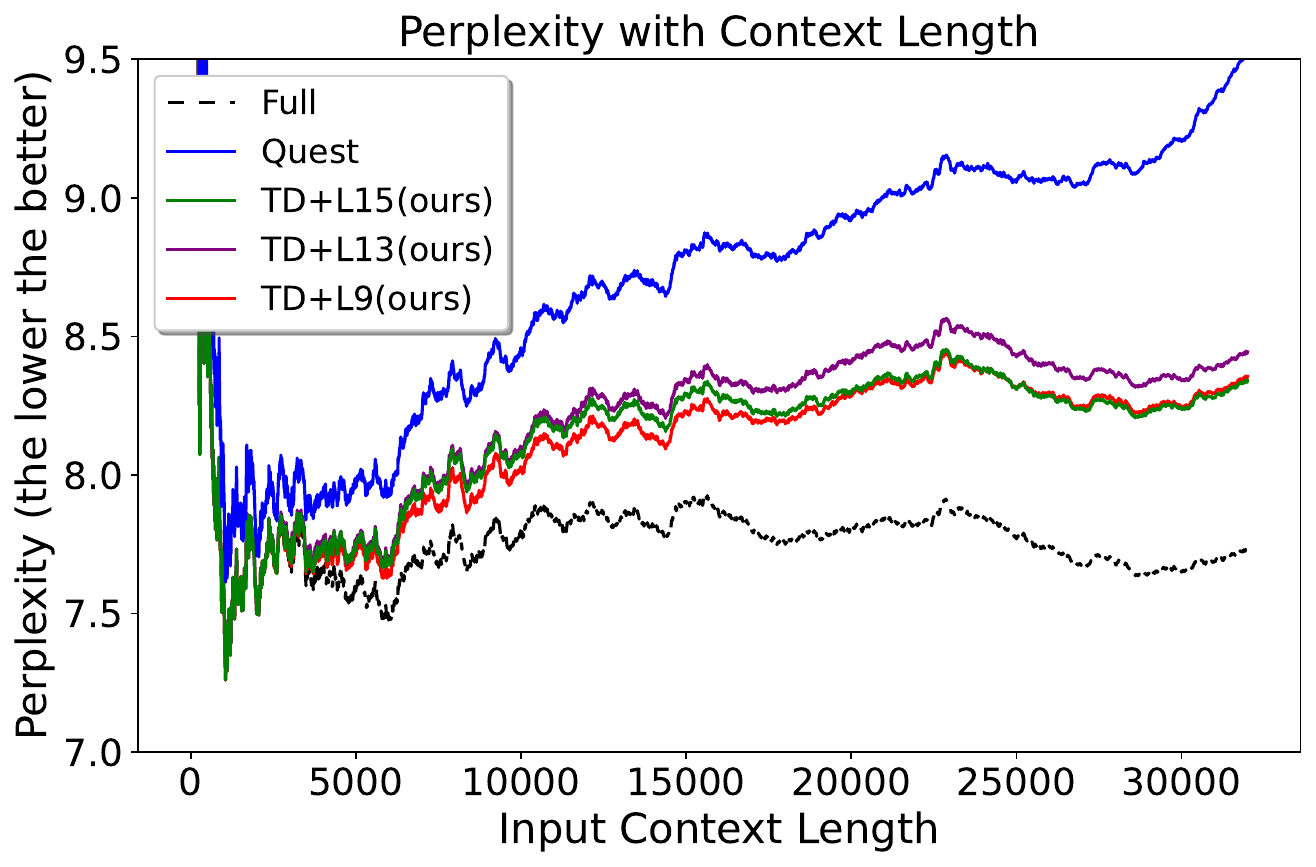}
        \caption{Token Budget = 2048}
        \label{fig:perplexity_2048}
    \end{subfigure}
    \hfill
    \begin{subfigure}[b]{0.49\textwidth}
        \centering
        \includegraphics[width=\textwidth]{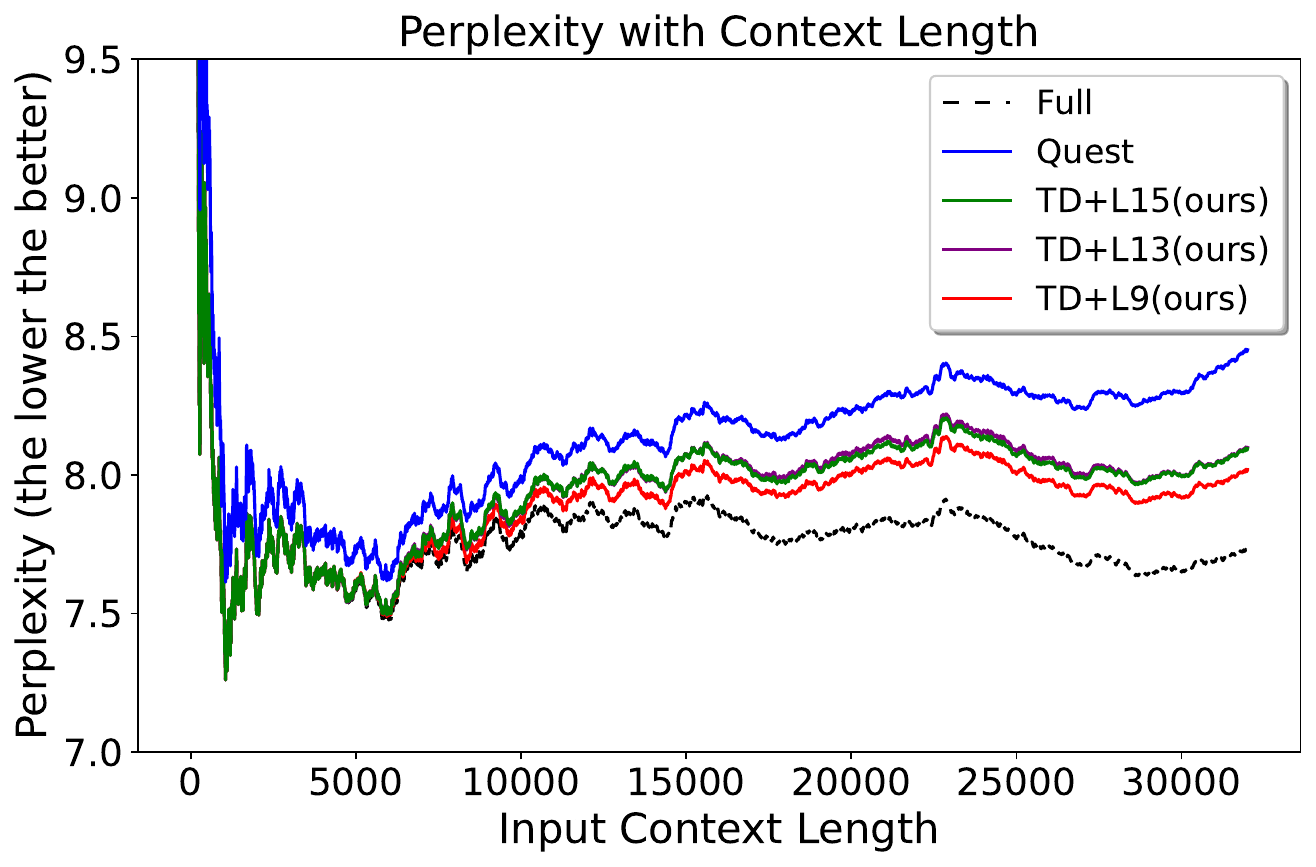}
        \caption{Token Budget = 4096}
        \label{fig:perplexity_4096}
    \end{subfigure}
    \caption{Perplexity evaluation on the PG-19 dataset from 0 to 32K tokens. The results compare \Sys{} with different token re-selection layers (L9, L13, L15) to Quest across token budgets (2048 \ref{fig:perplexity_2048}, 4096 \ref{fig:perplexity_4096}). Lower perplexity indicates better model performance. Full refers to dense attention as baseline.}
    \label{fig:perplexity_pg19_combined}
\end{figure}

Perplexity measures the negative likelihood of how well a model predicts the next word in a sequence, with lower values indicating better performance. We evaluate \Sys{} on Llama-3-8B-Instruct-Gradient-1048k with the PG-19 dataset, which includes up to 100 books, providing a comprehensive long-context benchmark.

As shown in Figure \ref{fig:perplexity_pg19_combined}, \Sys{}+L9/13/15 consistently achieves lower perplexity than Quest across all token budget options (2048, 4096). This indicates that \Sys{}'s position persistent sparse attention mechanism can effectively retain critical information without significantly sacrificing model accuracy, even as the sequence length grows, demonstrating its robustness for long-context inputs.

\subsubsection{LongBench Experiment}
\begin{table}[ht]
    \centering
    \caption{Performance comparison on eight LongBench datasets evaluating single/multi-document QA, summarization, and retrieval tasks using Llama-3-8B-Instruct-Gradient-1048k. \Sys{} outperforms Quest at a 4096 token budget and achieves an average score higher than full-weight attention. The maximum F1-score for each task is in bold.}
    \addtolength{\tabcolsep}{-1pt}    
    \begin{tabular}{lccccccccc}
        \toprule
        Method (K)/Task & MFQA & NrtQA & Qasp & 2Wiki & HotQA & QMSm & TrQA & PRe & Avg \\
        \midrule
        Full                   &   30.76   &   5.52   &   \textbf{14.56}   &   13.32   &   11.50   &   19.43   &   \textbf{86.56}   &   77.00   &   32.33 \\
        \midrule
        Quest \hfill(1024)           &   26.21   &   4.08   &   12.19   &   12.61   &   10.75   &   19.56   &   83.47   &   63.84   &   29.09 \\
        TD+L13 \hfill(1024)          &   28.57   &   \textbf{7.63}   &   11.11   &   13.56   &    9.82   &   \textbf{20.37}    &   79.78   &   75.17   & 30.75         \\
        \midrule
        Quest \hfill(4096)           &   28.92   &   3.74   &   13.63   &   12.83   &   12.15   &   19.36       &   85.91   &   72.50   & 31.13         \\
        TD+L13 \hfill(4096)          &   \textbf{30.94}   &   6.19   &   13.85   &   \textbf{14.40}   &   \textbf{13.71}   &   19.48   &   86.30   &   \textbf{78.00}   &   \textbf{32.86} \\
        \bottomrule
    \label{tab:longbench}
    \end{tabular}
    \addtolength{\tabcolsep}{1pt}    
\end{table}

We also evaluate \Sys{} on LongBench, a benchmark designed to test LLMs on long-context tasks across diverse NLP domains~\citep{bai2023longbench}. We focus on eight tasks: MultiFieldQA (MFQA), NarrativeQA (NrtQA), Qasper (Qasp), 2WikiMQA (2Wiki), HotpotQA (HotQA), QMSum (QMSm), TriviaQA (TrQA), and Passage Retrieval (PRe), which collectively composite a comprehensive evaluation benchmark in single/multi-document QA, summarization, and retrieval.

We evaluate all methods with LLaMA-3-8B-Instruct-Gradient-1048k. \Sys{} is compared against full-weight attention and Quest at token budgets of 1024 and 4096. As shown in \Cref{tab:longbench}, \Sys{} consistently outperforms Quest on all tasks at $K=4096$ and on five tasks at $K=1024$. Surprisingly, \Sys{}, in most cases, matches or exceeds full attention baseline with notable sparsity: 14\% on NrtQA, 50\% on MFQA, 80\% on Qasp, 50\% on 2WikiMQA, 32\% on HotQA, 29\% on QMSm, 35\% on TrQA, and 33\% on PRe. We hypothesize this is because \Sys{}'s token selection process can filter out irrelevant information, thus leading to higher performance.

These results demonstrate \Sys{}'s generic ability to select tokens with the highest attention scores, achieving competitive or superior performance while significantly reducing token usage, making it ideal for long-context scenarios.

\subsection{Efficiency Evaluation}
\label{sec:efficiency-eval}
\begin{figure}[htbp]
    \centering
    \includegraphics[width=0.98\textwidth]{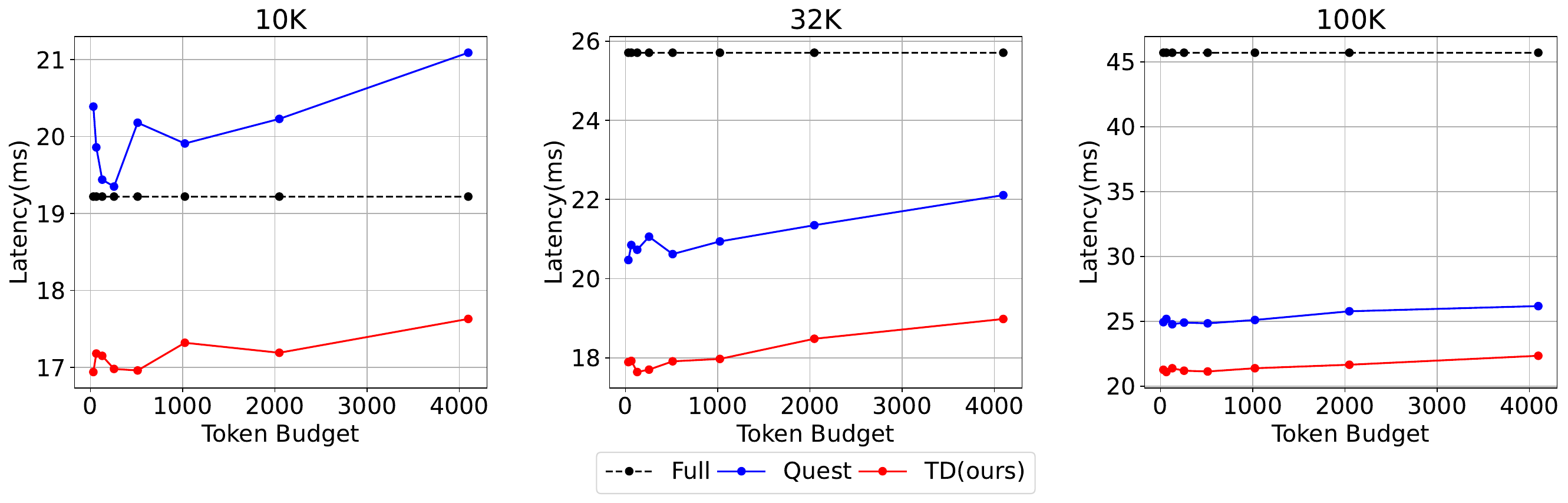}
    \caption{End-to-end latency results on LLaMA-2-7B model for Full attention baseline(Full), Quest, and \Sys{}(TD) when context length is 10K, 32K, and 100K, respectively.}
    \label{fig:e2e}
\end{figure}
To show the efficiency of \Sys{}, we write customized kernels for our approach and measure the end-to-end decoding latency. We conduct evaluation under the configuration of Llama-2-7B on one Nvidia A100 (80 GB HBM, SXM4) with CUDA 12.2. We compare \Sys{} with state-of-the-art full attention serving library FlashInfer~\citep{flashinfer} and also the Quest implementation. As shown in~\Cref{fig:e2e}, we can observe that \Sys{} can consistently outperform full attention baseline and Quest by a large margin under all token budgets and context lengths. \Sys{} achieves this through token pattern reuse to minimize the token selection overhead. Notice that the latest LLaMA-3 model shares the same architecture as LLaMA-2, except it uses Group-Query-Attention instead of Multi-Head-Attention. However, this does not affect the relative efficiency comparison against Quest and full attention.

\begin{figure}[ht]
    \centering
    \caption*{\includegraphics[width=.75\textwidth]{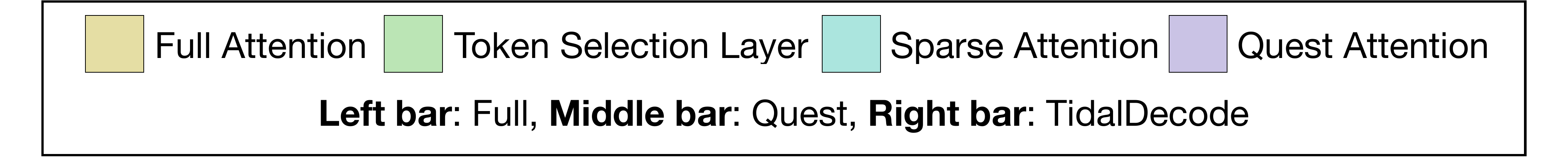}}
    \begin{subfigure}[b]{0.49\textwidth}
        \centering
        \includegraphics[width=\textwidth]{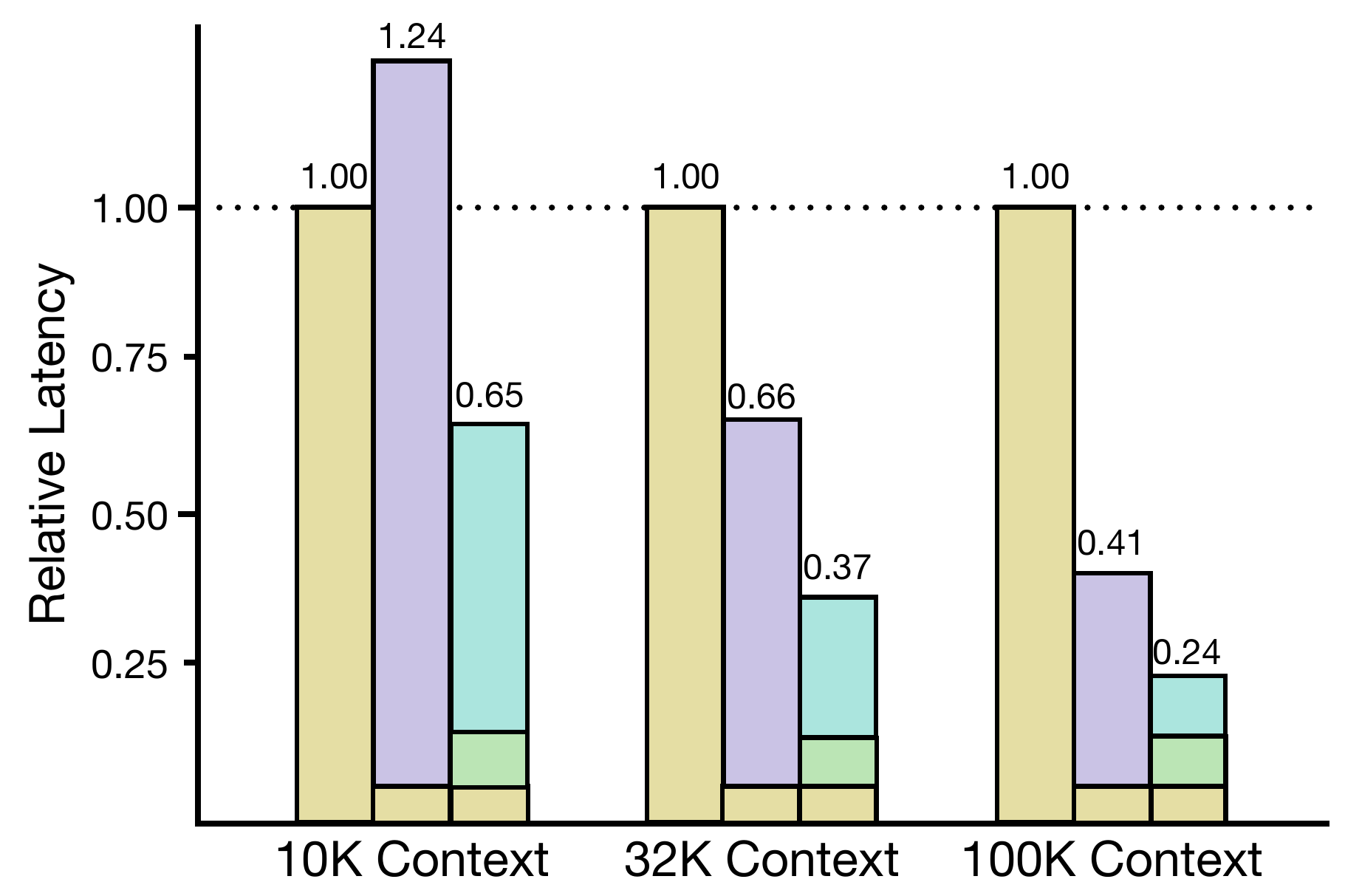}
        \caption{32 Layers}
        \label{fig:32-layer}
    \end{subfigure}
    \hfill
    \begin{subfigure}[b]{0.49\textwidth}
        \centering
        \includegraphics[width=\textwidth]{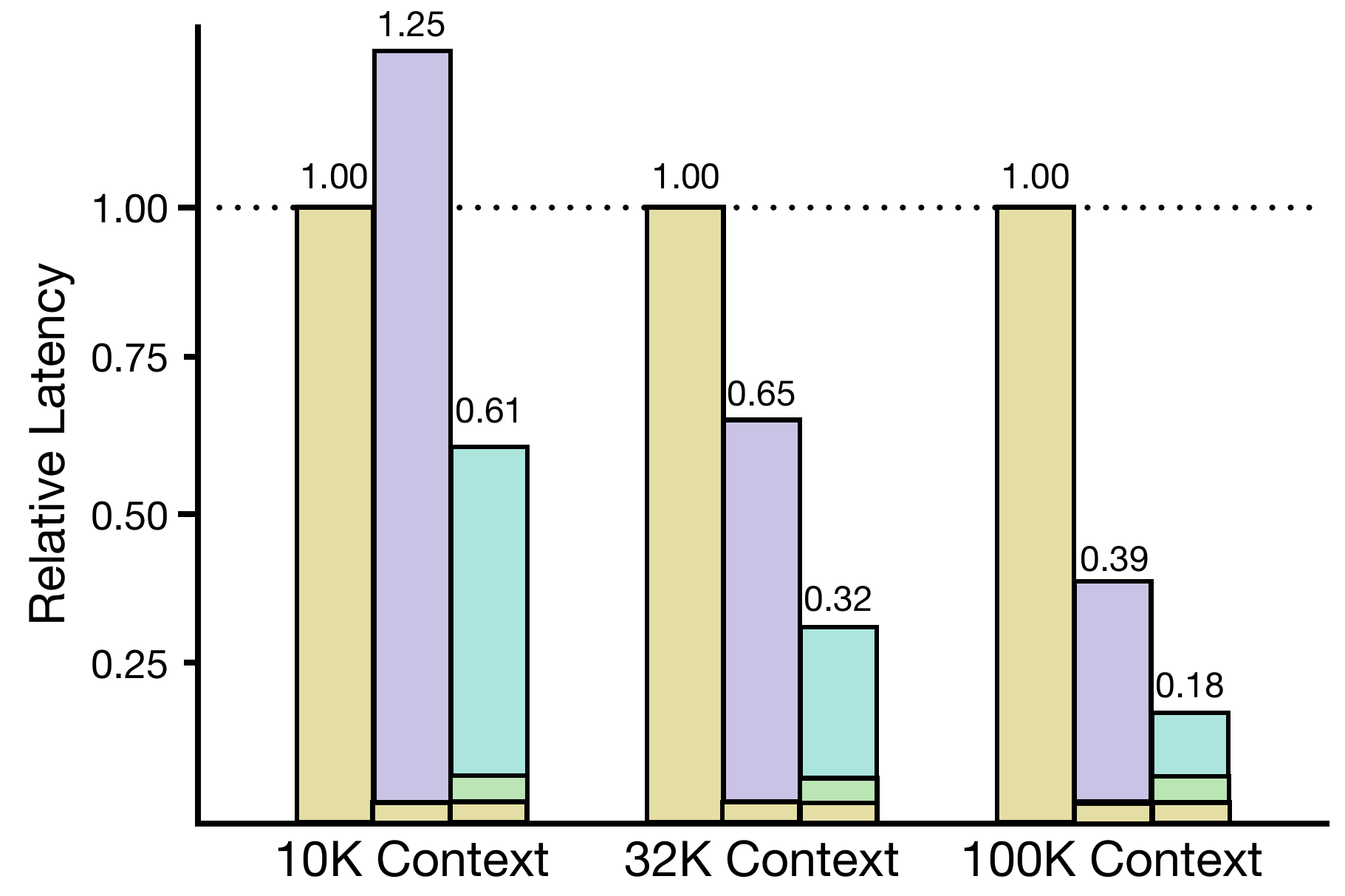}
        \caption{64 Layers}
        \label{fig:64-layer}
    \end{subfigure}
    \caption{Overall attention latency results for different methods on the LLaMA model with (a) 32 and (b) 64 layers. We use the full attention model as a reference and show \Sys{} and Quest's overall attention latency ratio. For each group of the bar plots, the left/middle/right bar denotes the full attention baseline, Quest, and \Sys{}, respectively. }
    \label{fig:kernel23}
\end{figure}

In~\Cref{fig:kernel23}, we compare the overall attention latency between different methods on the LLaMA model with 32/64 layers. For the 32-layer LLaMA model, we have 2 full attention layers + 2 token selection layers + 28 sparse attention layers, while Quest has 2 full attention layers + 30 Quest attention layers. For the 64-layer LLaMA model, we have 2 full attention layers + 2 token selection layers + 60 sparse attention layers, while Quest has 2 full attention layers + 62 Quest attention layers. Thus, by completely removing the token estimation overhead in the sparse attention layers, for the 32-layer and 64-layer LLaMA model under all context lengths, \Sys{} can consistently achieve the lowest serving latency while bringing up to $5.56\times$ speed-up ratio against the full attention baseline and $2.17\times$ speed-up ratio against Quest. When the context length is 10K, Quest has a higher latency due to the token selection overhead, which aligns with the end-to-end results in~\Cref{fig:e2e}. In contrast, \Sys{} still achieves significant speed-up by utilizing the position persistent sparse attention mechanism. 

\begin{figure}[ht]
    \centering
    \includegraphics[width=0.8\textwidth]{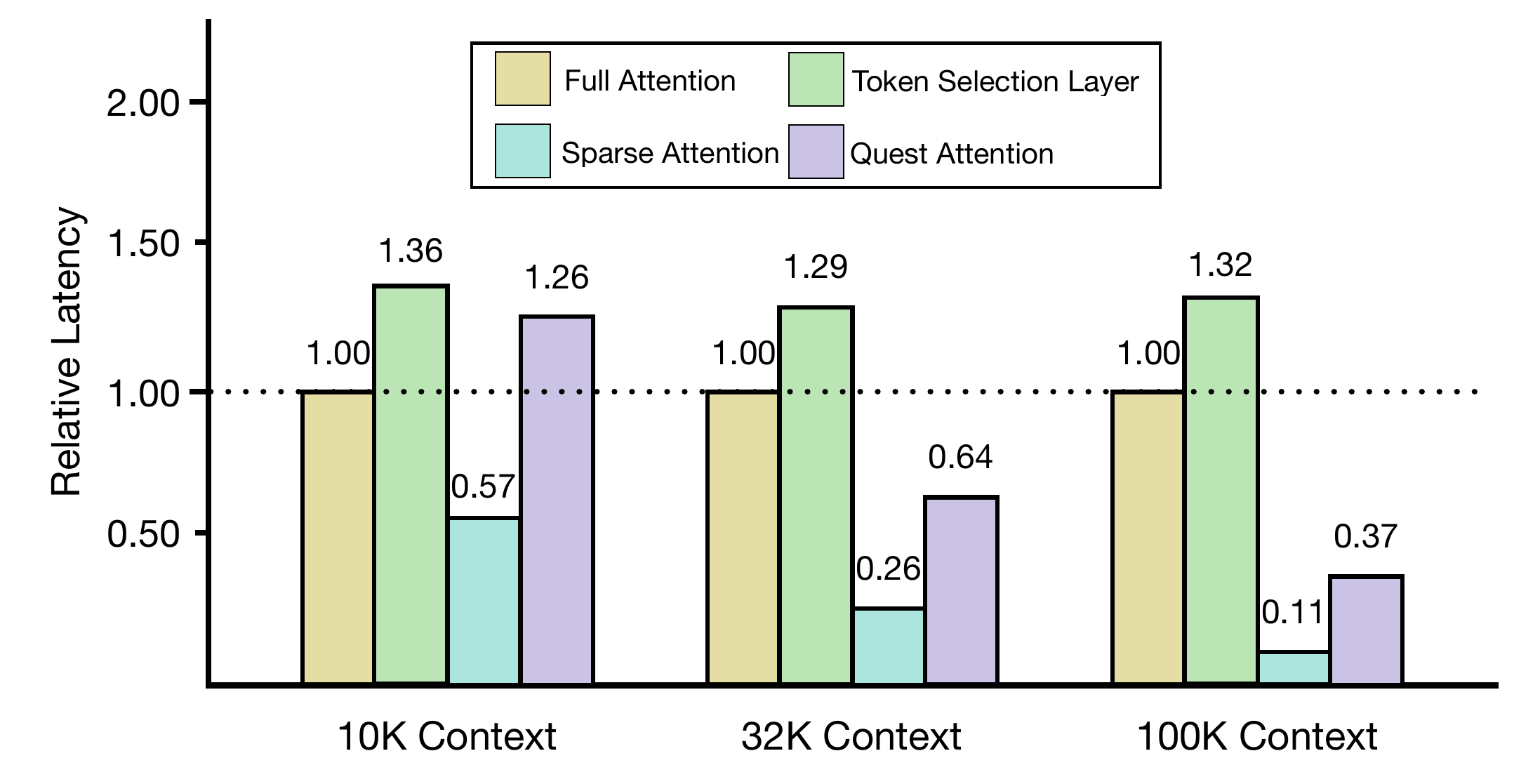}
    \caption{The breakdown latency results for the full attention, token selection attention, sparse attention, and Quest attention kernels over 10K, 32K, and 100K context length. We use full attention latency as a reference and report other kernels' relative latency ratio. We use a token budget of K=512 for \Sys{} and Quest across all evaluations.}
    \label{fig:kernel}
\end{figure}
In~\Cref{fig:kernel}, we further break down the latency comparison for different attention modules to show why \Sys{} can bring significant speed-up consistently. We compare different attention modules, namely, the full attention layer, the token selection layer, \Sys{}'s sparse attention layer, and the Quest attention layer over the 10K, 32K, and 100K context length. We can observe that, as \Sys{}'s sparse attention kernel can directly reuse previous token patterns, it completely removes the important token estimation overhead in the Quest attention kernel, resulting in up to $3.36\times$ speed-up compared with the Quest implementation. On the other hand, even though \Sys{}'s token selection layer has a slightly higher latency, we only have two token selection layers even in the 70B LLaMA model that has 64 layers in total. 
\subsection{Sensitivity analysis on token re-selection layer}
\label{sec:ablation-eval}
\begin{figure}[htp!]
    \centering
    \begin{subfigure}[b]{0.45\textwidth}
        \centering
        \includegraphics[width=\textwidth]{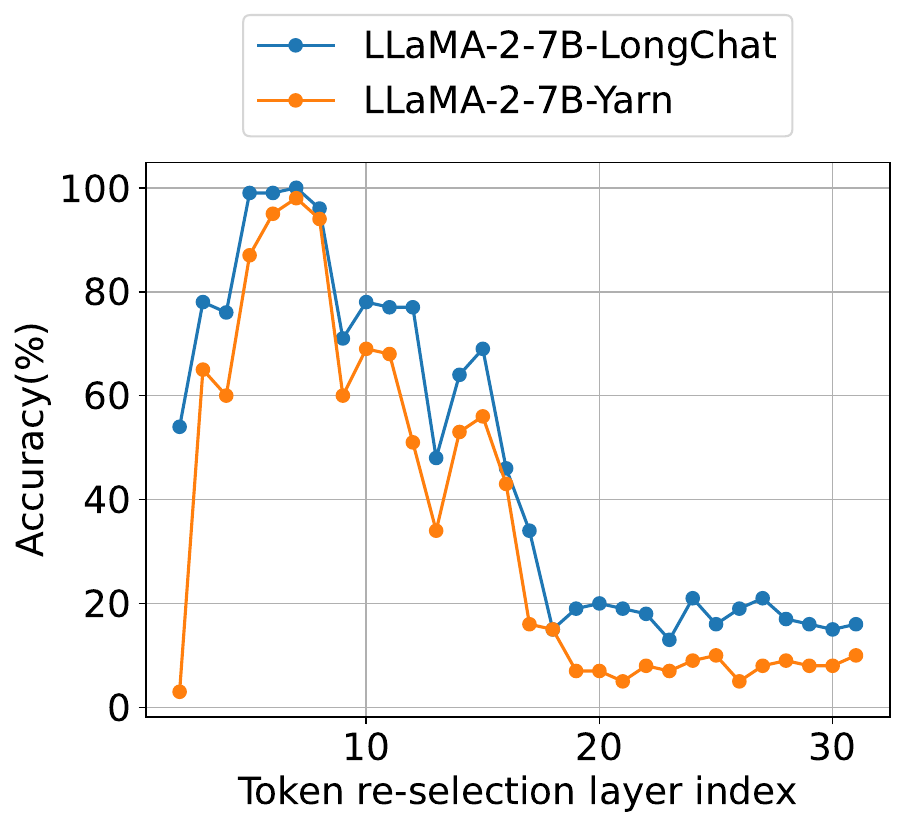}
        \caption{LLaMA-2 Model}
        \label{fig:llama-2-layer}
    \end{subfigure}
    \hfill
    \begin{subfigure}[b]{0.45\textwidth}
        \centering
        \includegraphics[width=\textwidth]{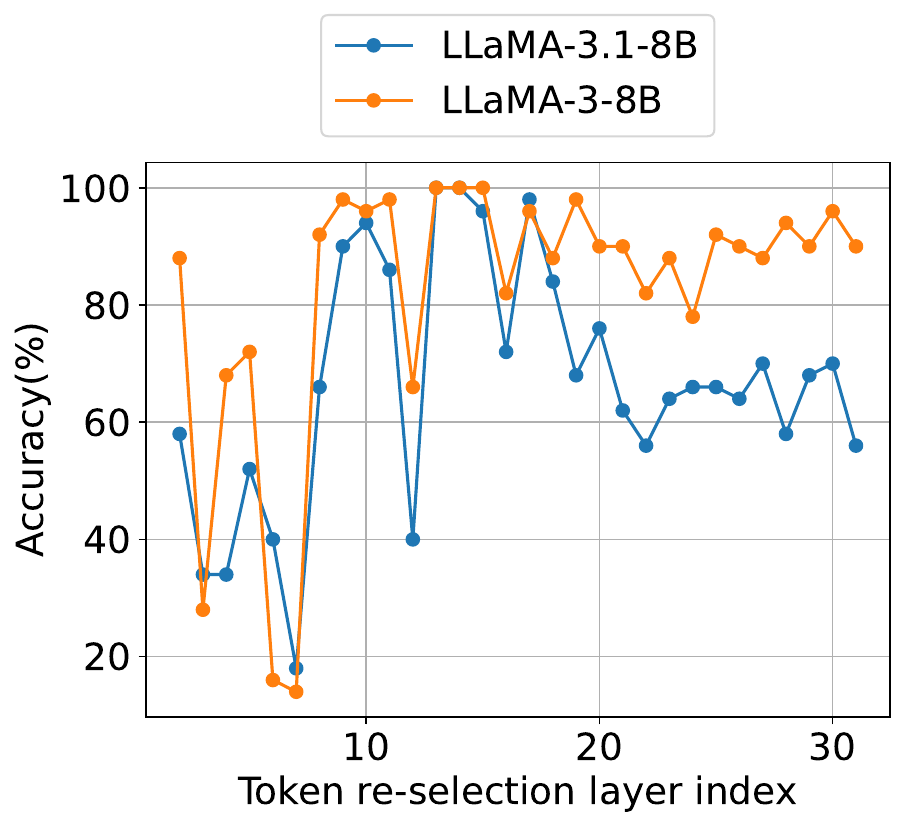}
        \caption{LLaMA-3 Model}
        \label{fig:llama-3-layer}
    \end{subfigure}
    \caption{Sensitivity study on the choice of different token re-selection layer. We evaluate LLaMA-2-7B-LongChat, LLaMA-2-7B-Yarn, LLaMA-3-8B, and LLaMA-3.1-8B with \Sys{} with a token budget of 256.}
    \label{fig:layer-ablation}
\end{figure}
In this section, we conduct sensitivity studies for different choices of the token re-selection layer. As \Sys{} only has one token re-selection layer in the middle, it is critical to choose the best-performed one. As shown in~\Cref{fig:layer-ablation}, we have two interesting findings: (1). Different choices of token re-selection layers can significantly affect the accuracy of the results (2). For models within the same model family, the optimal token re-selection layer is consistent over different tasks. In our setup, the optimal token re-selection layer for the LLaMA-2-7B model is layer 7, while for the LLaMA-3-8B/LLaMA-3.1-8B model is layer 13. A concurrent KV cache compression work also identifies that layer 13 is surprisingly important for their approach as well~\citep{shi2024discovering}. For a more detailed sensitivity results on the choice of different token re-selection layers, please refer to the appendix for more results.
\section{Conclusion}
\label{sec:conclusion}
To conclude, we introduce \Sys{}, an efficient LLM decoding framework with sparse attention. On observing the correlation of the pattern of tokens with the highest attention scores across different consecutive layers, \Sys{} proposes only to select tokens twice: once at the beginning layers and once in the middle layer to serve as a token re-selection layer. We find that using two token selection layers is necessary and sufficient to achieve high-generation quality. Additionally, by reusing the token patterns throughout the sparse attention layer, \Sys{} greatly reduces the token selection overhead, resulting in a significant end-to-end speed-up ratio against existing methods. More interestingly, the optimal choice of the token re-selection layer is consistent across different tasks if the model is in the same model family.

\section*{Acknowledgment}
This research is supported by NSF awards CNS-2147909, CNS-2211882, CNS-2239351, and research awards from Amazon, Cisco, Google, Meta, Oracle, Qualcomm, and Samsung. 


\bibliography{iclr2025_conference}

\begin{thebibliography}{29}
\providecommand{\natexlab}[1]{#1}
\providecommand{\url}[1]{\texttt{#1}}
\expandafter\ifx\csname urlstyle\endcsname\relax
  \providecommand{\doi}[1]{doi: #1}\else
  \providecommand{\doi}{doi: \begingroup \urlstyle{rm}\Url}\fi

\bibitem[AI(2024{\natexlab{a}})]{gradient-ai-llama-3-8B}
Gradient AI.
\newblock Llama-3-8b-instruct-gradient-1048k.
\newblock \url{https://huggingface.co/gradientai/Llama-3-8B-Instruct-Gradient-1048k}, 2024{\natexlab{a}}.
\newblock Accessed: 2024-09-26.

\bibitem[AI(2024{\natexlab{b}})]{llama3-1}
Meta AI.
\newblock Llama 3.1: Advanced open-source language model.
\newblock \url{https://ai.meta.com/blog/meta-llama-3-1/}, 2024{\natexlab{b}}.
\newblock Accessed: 2024-09-26.

\bibitem[Anthropic(2024)]{claude}
Anthropic.
\newblock Claude 3.5 sonnet.
\newblock \url{https://www.anthropic.com/news/claude-3-5-sonnet}, 2024.
\newblock [Accessed 20-06-2024].

\bibitem[Bai et~al.(2023)Bai, Lv, Zhang, Lyu, Tang, Huang, Du, Liu, Zeng, Hou, Dong, Tang, and Li]{bai2023longbench}
Yushi Bai, Xin Lv, Jiajie Zhang, Hongchang Lyu, Jiankai Tang, Zhidian Huang, Zhengxiao Du, Xiao Liu, Aohan Zeng, Lei Hou, Yuxiao Dong, Jie Tang, and Juanzi Li.
\newblock Longbench: A bilingual, multitask benchmark for long context understanding, 2023.

\bibitem[Child et~al.(2019)Child, Gray, Radford, and Sutskever]{child2019generating}
Rewon Child, Scott Gray, Alec Radford, and Ilya Sutskever.
\newblock Generating long sequences with sparse transformers.
\newblock \emph{arXiv preprint arXiv:1904.10509}, 2019.

\bibitem[Choromanski et~al.(2020)Choromanski, Likhosherstov, Dohan, Song, Gane, Sarlos, Hawkins, Davis, Mohiuddin, Kaiser, et~al.]{choromanski2020rethinking}
Krzysztof Choromanski, Valerii Likhosherstov, David Dohan, Xingyou Song, Andreea Gane, Tamas Sarlos, Peter Hawkins, Jared Davis, Afroz Mohiuddin, Lukasz Kaiser, et~al.
\newblock Rethinking attention with performers.
\newblock \emph{arXiv preprint arXiv:2009.14794}, 2020.

\bibitem[Dao(2023)]{dao2023flashattention}
Tri Dao.
\newblock Flashattention-2: Faster attention with better parallelism and work partitioning.
\newblock \emph{arXiv preprint arXiv:2307.08691}, 2023.

\bibitem[Dao et~al.(2022)Dao, Fu, Ermon, Rudra, and Ré]{dao2022flashattention}
Tri Dao, Daniel~Y. Fu, Stefano Ermon, Atri Rudra, and Christopher Ré.
\newblock Flashattention: Fast and memory-efficient exact attention with io-awareness, 2022.

\bibitem[Ge et~al.(2024)Ge, Zhang, Liu, Zhang, Han, and Gao]{ge2024model}
Suyu Ge, Yunan Zhang, Liyuan Liu, Minjia Zhang, Jiawei Han, and Jianfeng Gao.
\newblock Model tells you what to discard: Adaptive kv cache compression for llms, 2024.

\bibitem[Huang et~al.(2021)Huang, Cao, Parulian, Ji, and Wang]{huang-etal-2021-efficient}
Luyang Huang, Shuyang Cao, Nikolaus Parulian, Heng Ji, and Lu~Wang.
\newblock Efficient attentions for long document summarization.
\newblock In \emph{Proceedings of the 2021 Conference of the North American Chapter of the Association for Computational Linguistics: Human Language Technologies}, pp.\  1419--1436, Online, June 2021. Association for Computational Linguistics.
\newblock \doi{10.18653/v1/2021.naacl-main.112}.
\newblock URL \url{https://aclanthology.org/2021.naacl-main.112}.

\bibitem[Kitaev et~al.(2020)Kitaev, Kaiser, and Levskaya]{kitaev2020reformer}
Nikita Kitaev, {\L}ukasz Kaiser, and Anselm Levskaya.
\newblock Reformer: The efficient transformer.
\newblock \emph{arXiv preprint arXiv:2001.04451}, 2020.

\bibitem[Kwon et~al.(2023)Kwon, Li, Zhuang, Sheng, Zheng, Yu, Gonzalez, Zhang, and Stoica]{kwon2023efficient}
Woosuk Kwon, Zhuohan Li, Siyuan Zhuang, Ying Sheng, Lianmin Zheng, Cody~Hao Yu, Joseph~E. Gonzalez, Hao Zhang, and Ion Stoica.
\newblock Efficient memory management for large language model serving with pagedattention.
\newblock In \emph{Proceedings of the ACM SIGOPS 29th Symposium on Operating Systems Principles}, 2023.

\bibitem[OpenAI(2024)]{gpt4}
OpenAI.
\newblock Introducing gpt-4o: our fastest and most affordable flagship model.
\newblock \url{https://platform.openai.com/docs/models}, 2024.
\newblock [Accessed 28-05-2024].

\bibitem[Peng et~al.(2023)Peng, Quesnelle, Fan, and Shippole]{peng2023yarn}
Bowen Peng, Jeffrey Quesnelle, Honglu Fan, and Enrico Shippole.
\newblock Yarn: Efficient context window extension of large language models, 2023.

\bibitem[Rae et~al.(2019)Rae, Potapenko, Jayakumar, Hillier, and Lillicrap]{raecompressive2019}
Jack~W Rae, Anna Potapenko, Siddhant~M Jayakumar, Chloe Hillier, and Timothy~P Lillicrap.
\newblock Compressive transformers for long-range sequence modelling.
\newblock \emph{arXiv preprint}, 2019.
\newblock URL \url{https://arxiv.org/abs/1911.05507}.

\bibitem[Ram et~al.(2023)Ram, Levine, Dalmedigos, Muhlgay, Shashua, Leyton-Brown, and Shoham]{ram2023incontextretrievalaugmentedlanguagemodels}
Ori Ram, Yoav Levine, Itay Dalmedigos, Dor Muhlgay, Amnon Shashua, Kevin Leyton-Brown, and Yoav Shoham.
\newblock In-context retrieval-augmented language models, 2023.
\newblock URL \url{https://arxiv.org/abs/2302.00083}.

\bibitem[Ribar et~al.(2023)Ribar, Chelombiev, Hudlass-Galley, Blake, Luschi, and Orr]{ribar2023sparq}
Luka Ribar, Ivan Chelombiev, Luke Hudlass-Galley, Charlie Blake, Carlo Luschi, and Douglas Orr.
\newblock Sparq attention: Bandwidth-efficient llm inference, 2023.

\bibitem[Shi et~al.(2024)Shi, Ming, Nguyen, Liang, and Joty]{shi2024discovering}
Zhenmei Shi, Yifei Ming, Xuan-Phi Nguyen, Yingyu Liang, and Shafiq Joty.
\newblock Discovering the gems in early layers: Accelerating long-context llms with 1000x input token reduction.
\newblock \emph{arXiv preprint arXiv:2409.17422}, 2024.

\bibitem[Su et~al.(2023)Su, Lu, Pan, Murtadha, Wen, and Liu]{su2023roformer}
Jianlin Su, Yu~Lu, Shengfeng Pan, Ahmed Murtadha, Bo~Wen, and Yunfeng Liu.
\newblock Roformer: Enhanced transformer with rotary position embedding, 2023.

\bibitem[Tang et~al.(2024)Tang, Zhao, Zhu, Xiao, Kasikci, and Han]{tang2024questqueryawaresparsityefficient}
Jiaming Tang, Yilong Zhao, Kan Zhu, Guangxuan Xiao, Baris Kasikci, and Song Han.
\newblock Quest: Query-aware sparsity for efficient long-context llm inference, 2024.
\newblock URL \url{https://arxiv.org/abs/2406.10774}.

\bibitem[Touvron et~al.(2023)Touvron, Lavril, Izacard, Martinet, Lachaux, Lacroix, Rozière, Goyal, Hambro, Azhar, Rodriguez, Joulin, Grave, and Lample]{touvron2023llama}
Hugo Touvron, Thibaut Lavril, Gautier Izacard, Xavier Martinet, Marie-Anne Lachaux, Timothée Lacroix, Baptiste Rozière, Naman Goyal, Eric Hambro, Faisal Azhar, Aurelien Rodriguez, Armand Joulin, Edouard Grave, and Guillaume Lample.
\newblock Llama: Open and efficient foundation language models, 2023.

\bibitem[Vaswani et~al.(2023)Vaswani, Shazeer, Parmar, Uszkoreit, Jones, Gomez, Kaiser, and Polosukhin]{vaswani2023attentionneed}
Ashish Vaswani, Noam Shazeer, Niki Parmar, Jakob Uszkoreit, Llion Jones, Aidan~N. Gomez, Lukasz Kaiser, and Illia Polosukhin.
\newblock Attention is all you need, 2023.
\newblock URL \url{https://arxiv.org/abs/1706.03762}.

\bibitem[Wang et~al.(2024)Wang, Salmani, Omidi, Ren, Rezagholizadeh, and Eshaghi]{wang2024limitssurveytechniquesextend}
Xindi Wang, Mahsa Salmani, Parsa Omidi, Xiangyu Ren, Mehdi Rezagholizadeh, and Armaghan Eshaghi.
\newblock Beyond the limits: A survey of techniques to extend the context length in large language models, 2024.
\newblock URL \url{https://arxiv.org/abs/2402.02244}.

\bibitem[Wei et~al.(2023)Wei, Wang, Schuurmans, Bosma, Ichter, Xia, Chi, Le, and Zhou]{wei2023chainofthoughtpromptingelicitsreasoning}
Jason Wei, Xuezhi Wang, Dale Schuurmans, Maarten Bosma, Brian Ichter, Fei Xia, Ed~Chi, Quoc Le, and Denny Zhou.
\newblock Chain-of-thought prompting elicits reasoning in large language models, 2023.
\newblock URL \url{https://arxiv.org/abs/2201.11903}.

\bibitem[Xiao et~al.(2023)Xiao, Tian, Chen, Han, and Lewis]{xiao2023streamingllm}
Guangxuan Xiao, Yuandong Tian, Beidi Chen, Song Han, and Mike Lewis.
\newblock Efficient streaming language models with attention sinks.
\newblock \emph{arXiv}, 2023.

\bibitem[Ye et~al.(2024)Ye, Lai, Lu, Lin, Zheng, Chen, Chen, and Ceze]{flashinfer}
Zihao Ye, Ruihang Lai, Roy Lu, Chien-Yu Lin, Size Zheng, Lequn Chen, Tianqi Chen, and Luis Ceze.
\newblock Cascade inference: Memory bandwidth efficient shared prefix batch decoding.
\newblock \url{https://flashinfer.ai/2024/01/08/cascade-inference.html}, Jan 2024.
\newblock URL \url{https://flashinfer.ai/2024/01/08/cascade-inference.html}.
\newblock Accessed on 2024-02-01.

\bibitem[Zhang et~al.(2023)Zhang, Sheng, Zhou, Chen, Zheng, Cai, Song, Tian, Ré, Barrett, Wang, and Chen]{zhang2023h2o}
Zhenyu Zhang, Ying Sheng, Tianyi Zhou, Tianlong Chen, Lianmin Zheng, Ruisi Cai, Zhao Song, Yuandong Tian, Christopher Ré, Clark Barrett, Zhangyang Wang, and Beidi Chen.
\newblock H$_2$o: Heavy-hitter oracle for efficient generative inference of large language models, 2023.

\bibitem[Zhang et~al.(2024{\natexlab{a}})Zhang, Sheng, Zhou, Chen, Zheng, Cai, Song, Tian, R{\'e}, Barrett, et~al.]{zhang2024h2o}
Zhenyu Zhang, Ying Sheng, Tianyi Zhou, Tianlong Chen, Lianmin Zheng, Ruisi Cai, Zhao Song, Yuandong Tian, Christopher R{\'e}, Clark Barrett, et~al.
\newblock H2o: Heavy-hitter oracle for efficient generative inference of large language models.
\newblock \emph{Advances in Neural Information Processing Systems}, 36, 2024{\natexlab{a}}.

\bibitem[Zhang et~al.(2024{\natexlab{b}})Zhang, Zhu, Yang, Xu, Li, Phothilimthana, and Jia]{pmlr-v235-zhang24cq}
Zhihao Zhang, Alan Zhu, Lijie Yang, Yihua Xu, Lanting Li, Phitchaya~Mangpo Phothilimthana, and Zhihao Jia.
\newblock Accelerating iterative retrieval-augmented language model serving with speculation.
\newblock In Ruslan Salakhutdinov, Zico Kolter, Katherine Heller, Adrian Weller, Nuria Oliver, Jonathan Scarlett, and Felix Berkenkamp (eds.), \emph{Proceedings of the 41st International Conference on Machine Learning}, volume 235 of \emph{Proceedings of Machine Learning Research}, pp.\  60626--60643. PMLR, 21--27 Jul 2024{\natexlab{b}}.
\newblock URL \url{https://proceedings.mlr.press/v235/zhang24cq.html}.

\end{thebibliography}
\bibliographystyle{iclr2025_conference}

\newpage
\appendix
\section{Appendix}
\subsection{LongBench for LLaMA-3.1-8B-Instruct}
\label{app:longbench}
\begin{table}[ht]
\centering
\caption{Performance comparison on eight LongBench datasets evaluating single/multi-document QA, summarization, and retrieval tasks using Llama-3.1-8B-Instruct. The maximum F1-score for each task is in bold.}
\addtolength{\tabcolsep}{-1pt} 
\begin{tabular}{l|c|c|c|c|c|c|c|c|c}
\toprule
\textbf{Method} & \textbf{MFQA} & \textbf{NrtQA} & \textbf{Qasp} & \textbf{2Wiki} & \textbf{HotQA} & \textbf{QSm} & \textbf{TrQA} & \textbf{Pre} & \textbf{Avg} \\
\midrule
Full & \textbf{27.02} & 25.59 & \textbf{13.05} & 16.64 & \textbf{16.86} & \textbf{23.88} & 91.48 & \textbf{97.67} & \textbf{39.02} \\
Quest (1024) & 22.35 & 14.89 & 12.44 & 14.24 & 14.12 & 23.86 & 81.71 & 95.73 & 34.92 \\
TD+13 (1024) & 23.70 & 23.25 & 11.14 & 13.53 & 13.72 & 22.69 & \textbf{92.35} & 92.15 & 36.57 \\
Quest (4096) & 26.34 & 21.17 & 11.99 & 15.61 & 16.26 & 23.61 & 90.73 & 96.35 & 37.76 \\
TD+13 (4096) & 25.89 & \textbf{26.29} & 12.65 & \textbf{16.86} & 15.94 & 23.27 & 90.22 & 95.47 & 38.32 \\
\bottomrule
\end{tabular}
\addtolength{\tabcolsep}{1pt} 
\end{table}
\Sys{} and full-weight attention share the maximum F1 scores for all tasks, achieving the best scores in three tasks (NrtQA, 2Wiki, and TrQA). \Sys{} significantly outperforms Quest in 4/8 tasks (NrtQA, Qasp, 2Wiki, and TrQA) and full-attention in 3/8 tasks (NrtQA, 2Wiki, and TrQA). For other tasks, we stay close to the full attention and also obtains a higher average score than Quest.

\subsection{End-to-end efficiency evaluation results}
\label{app:e2e}
\begin{table}[H]
\centering
\caption{\Sys{} end-to-end efficiency results on LLaMA-2-7B}
\addtolength{\tabcolsep}{-4pt} 
\begin{tabular}{cccccccccc}
\toprule
\multirow{2}{*}{Context Length} & \multirow{2}{*}{\begin{tabular}[c]{@{}c@{}}Full Attention\\ (ms)\end{tabular}} & \multicolumn{8}{c}{TidalDecode(ms)}                              \\
                                &                                                                                & K=32  & K=64  & K=128 & K=256 & K=512 & K=1024 & K=2048 & K=4096 \\ \midrule
10K                             & 19.22                                                                          & 16.94 & 17.18 & 17.15 & 16.98 & 16.96 & 17.32  & 17.19  & 17.63  \\
32K                             & 25.71                                                                          & 17.89 & 17.92 & 17.64 & 17.70 & 17.91 & 17.97  & 18.48  & 18.98  \\
100K                            & 45.70                                                                          & 21.26 & 21.09 & 21.38 & 21.19 & 21.13 & 21.38  & 21.65  & 22.34  \\ \bottomrule
\end{tabular}
\addtolength{\tabcolsep}{4pt} 
\end{table}
\begin{table}[H]
\centering
\caption{Quest end-to-end efficiency results on LLaMA-2-7B}
\addtolength{\tabcolsep}{-4pt} 
\begin{tabular}{cccccccccc}
\toprule
\multirow{2}{*}{Context Length} & \multirow{2}{*}{\begin{tabular}[c]{@{}c@{}}Full Attention\\ (ms)\end{tabular}} & \multicolumn{8}{c}{TidalDecode(ms)}                              \\
                                &                                                                                & K=32  & K=64  & K=128 & K=256 & K=512 & K=1024 & K=2048 & K=4096 \\ \midrule
10K                             & 19.22                                                                          & 20.39 & 19.86 & 19.44 & 19.35 & 20.18 & 19.91 & 20.23 & 21.09  \\
32K                             & 25.71                                                                          & 20.47 & 20.85 & 20.73 & 21.06 & 20.62 & 20.94 & 21.35 & 22.11  \\
100K                            & 45.70                                                                          & 24.93 & 25.18 & 24.77 & 24.90 & 24.84 & 25.10 & 25.77 & 26.17  \\ \bottomrule
\end{tabular}
\addtolength{\tabcolsep}{4pt} 
\end{table}
\newpage
\subsection{Full sensitivity studies on different token re-selection layer}
\begin{table}[ht]
\centering
\caption{Sensitivity study of re-selection layer (RL) on 10k-context-length Needle-in-the-Haystack test for LLaMA-3.2-3B-Instruct with \Sys{}. The best accuracy for each token budget (K) is in bold.}
\begin{tabular}{c|c|c|c|c|c}
\hline
\textbf{RL/K} & 32 & 64 & 128 & 256 & 512 \\
\hline
\Sys{}+L2  & 2\% & 12\% & 14\% & 16\% & 30\% \\
\Sys{}+L3  & 2\% & 6\% & 8\% & 10\% & 24\% \\
\Sys{}+L4  & 6\% & 14\% & 16\% & 20\% & 28\% \\
\Sys{}+L5  & 2\% & 10\% & 22\% & 26\% & 36\% \\
\Sys{}+L6  & 18\% & 26\% & 26\% & 32\% & 46\% \\
\Sys{}+L7  & 6\% & 10\% & 16\% & 18\% & 28\% \\
\Sys{}+L8  & 20\% & 20\% & 46\% & 60\% & 84\% \\
\Sys{}+L9  & 6\% & 12\% & 32\% & 58\% & 66\% \\
\Sys{}+L10 & 44\% & 58\% & 50\% & 60\% & 64\% \\
\Sys{}+L11 & 4\% & 12\% & 16\% & 22\% & 28\% \\
\Sys{}+L12 & \textbf{50\%} & \textbf{84\%} & \textbf{96\%} & \textbf{98\%} & 98\% \\
\Sys{}+L13 & 42\% & 80\% & 94\% & \textbf{98\% }& \textbf{100\%} \\
\Sys{}+L14 & 28\% & 44\% & 54\% & 60\% & 72\% \\
\Sys{}+L15 & 2\% & 8\% & 16\% & 22\% & 36\% \\
\Sys{}+L16 & 4\% & 16\% & 12\% & 22\% & 34\% \\
\Sys{}+L17 & 2\% & 6\% & 16\% & 20\% & 32\% \\
\Sys{}+L18 & 2\% & 10\% & 12\% & 18\% & 28\% \\
\Sys{}+L19 & 2\% & 6\% & 10\% & 18\% & 32\% \\
\Sys{}+L20 & 6\% & 10\% & 12\% & 18\% & 24\% \\
\Sys{}+L21 & 6\% & 8\% & 10\% & 16\% & 26\% \\
\Sys{}+L22 & 6\% & 0\% & 12\% & 12\% & 26\% \\
\Sys{}+L23 & 4\% & 14\% & 14\% & 18\% & 26\% \\
\Sys{}+L24 & 2\% & 10\% & 16\% & 20\% & 28\% \\
\Sys{}+L25 & 4\% & 8\% & 14\% & 16\% & 22\% \\
\Sys{}+L26 & 2\% & 10\% & 10\% & 22\% & 26\% \\
\Sys{}+L27 & 0\% & 10\% & 12\% & 22\% & 26\% \\
\hline
Quest & 46\% & 56\% & 72\% & 88\% & 96\% \\ 
\hline
\end{tabular}
\end{table}

\begin{table}[ht]
\centering
\caption{Sensitivity study of re-selection layer (RL) on 10k-context-length Needle-in-the-Haystack test for LLaMA-3.1-8B-Instruct with \Sys{}. The best accuracy for each token budget (K) is in bold. Layer 13 and Layer 14 are the best two re-selection layers for accuracy. }
\begin{tabular}{c|c|c|c|c}
\hline
\textbf{RL/K} & 32 & 64 & 128 & 256 \\
\hline
\Sys{}+L2  & 36\% & 38\%& 46\%& 58\%\\
\Sys{}+L3  & 8\% & 10\%& 14\%& 34\%\\
\Sys{}+L4  & 0\% & 10\%& 16\%& 34\%\\
\Sys{}+L5  & 14\%& 30\%& 52\%& 52\%\\
\Sys{}+L6  & 8\%& 12\%& 28\%& 40\%\\
\Sys{}+L7  & 6\%& 10\%& 10\%& 18\%\\
\Sys{}+L8  & 34\%& 44\%& 50\%& 66\%\\
\Sys{}+L9  & 64\%& 78\%& 82\%& 90\%\\
\Sys{}+L10 & 56\%& 74\%& 84\%& 94\%\\
\Sys{}+L11 & 52\%& 76\%& 82\%& 86\%\\
\Sys{}+L12 & 8\%& 10\%& 28\%& 40\%\\
\Sys{}+L13 & \textbf{100\%}& \textbf{100\%}& \textbf{100\%}& \textbf{100\%}\\
\Sys{}+L14 & 98\%& \textbf{100\%}& \textbf{100\%}& \textbf{100\%}\\
\Sys{}+L15 & 56\%& 78\%& 88\%& 96\%\\
\Sys{}+L16 & 18\%& 46\%& 54\%& 72\%\\
\Sys{}+L17 & 64\%& 74\%& 86\%& 98\%\\
\Sys{}+L18 & 64\%& 70\%& 74\%& 84\%\\
\Sys{}+L19 & 58\%& 50\%& 60\%& 68\%\\
\Sys{}+L20 & 68\%& 60\%& 62\%& 76\%\\
\Sys{}+L21 & 40\%& 48\%& 48\%& 62\%\\
\Sys{}+L22 & 28\%& 38\%& 46\%& 56\%\\
\Sys{}+L23 & 40\%& 46\%& 52\%& 64\%\\
\Sys{}+L24 & 30\%& 46\%& 54\%& 66\%\\
\Sys{}+L25 & 40\%& 54\%& 50\%& 66\%\\
\Sys{}+L26 & 34\%& 48\%& 62\%& 64\%\\
\Sys{}+L27 & 38\%& 50\%& 54\%& 70\%\\
\Sys{}+L28 & 30\%& 40\%& 56\%& 58\%\\
\Sys{}+L29 & 32\%& 48\%& 56\%& 68\%\\
\Sys{}+L30 & 36\%& 48\%& 52\%& 70\%\\
\Sys{}+L31 & 30\%& 36\%& 42\%& 56\%\\
\hline
\end{tabular}
\end{table}

\begin{table}[htp]
\centering
\caption{Sensitivity study of re-selection layer (RL) on 10k-context-length Needle-in-the-Haystack test for LLaMA-3-70B-Instruct-Gradient-1048k; we first run top\_k = 512 and filter out those layers that do not achieve full accuracy with \Sys{}. The best accuracy for each token budget (K) is in bold. Layer 14 and Layer 31 are the best two Re-selection layers for accuracy.}
\begin{minipage}{.5\linewidth}
\centering
\begin{tabular}{c|c|c|c|c|c}
\hline
\textbf{RL/K} & 32 & 64 & 128 & 256 & 512 \\
\hline
L2  & - & - & - & - & 6\% \\
L3  & - & - & - & - & 37\% \\
L4  & - & - & - & - & 23\% \\
L5  & - & - & - & - & 63\% \\
L6  & - & - & - & - & 70\% \\
L7  & - & - & - & - & 90\% \\
L8  & - & - & - & - & 70\% \\
L9  & - & - & - & - & 30\% \\
L10  & - & - & - & - & 83\% \\
L11  & - & - & - & - & 70\% \\
L12  & - & - & - & - & 63\% \\
L13  & - & - & - & - & 50\% \\
L14  & 87\% & 93\% & \textbf{100\%} & \textbf{100\%} & \textbf{100\%} \\
L15  & - & - & - & - & 97\% \\
L16  & - & - & - & - & 63\% \\
L17  & - & - & - & - & 87\% \\
L18  & 50\% & 70\% & 83\% & 97\% & \textbf{100\%} \\
L19  & - & - & - & - & 93\% \\
L20  & - & - & - & - & 87\% \\
L21  & 53\% & 80\% & 93\% & 97\% & \textbf{100\%} \\
L22  & - & - & - & - & 97\% \\
L23  & 53\% & 93\% & 97\% & \textbf{100\%} & \textbf{100\%} \\
L24  & 33\% & 60\% & 77\% & 93\% & \textbf{100\%} \\
L25  & - & - & - & - & 80\% \\
L26  & - & - & - & - & 87\% \\
L27  & 50\% & 87\% & 93\% & 93\% & \textbf{100\%} \\
L28  & 80\% & 83\% & 93\% & 87\% & \textbf{100\%} \\
L29  & - & - & - & - & 97\% \\
L30  & 33\% & 67\% & 80\% & 90\% & \textbf{100\%} \\
L31  & \textbf{90\%} & \textbf{97\%} & \textbf{100\%} & \textbf{100\%} & \textbf{100\%} \\
L32  & 27\% & 73\% & 80\% & 97\% & \textbf{100\%} \\
\hline
\end{tabular}
\end{minipage}%
\begin{minipage}{.5\linewidth}
\centering
\begin{tabular}{c|c|c|c|c|c}
\hline
\textbf{RL/K} & 32 & 64 & 128 & 256 & 512 \\
\hline
L33  & 50\% & 87\% & 90\% & 93\% & \textbf{100\%} \\
L34  & - & - & - & - & 97\% \\
L35  & 70\% & 83\% & \textbf{100\%} & \textbf{100\%} & \textbf{100\%} \\
L36  & 50\% & 83\% & 97\% & 97\% & \textbf{100\%} \\
L37  & - & - & - & - & 90\% \\
L38  & 37\% & 83\% & 83\% & 80\% & \textbf{100\%} \\
L39  & - & - & - & - & 87\% \\
L40  & - & - & - & - & 50\% \\
L41  & - & - & - & - & 97\% \\
L42  & - & - & - & - & 53\% \\
L43  & - & - & - & - & 67\% \\
L44  & - & - & - & - & 83\% \\
L45  & - & - & - & - & 70\% \\
L46  & - & - & - & - & 63\% \\
L47  & - & - & - & - & 77\% \\
L48  & - & - & - & - & 97\% \\
L49  & - & - & - & - & 77\% \\
L50  & - & - & - & - & 70\% \\
L51  & - & - & - & - & 93\% \\
L52  & - & - & - & - & 77\% \\
L53  & - & - & - & - & 70\% \\
L54  & - & - & - & - & 60\% \\
L55  & - & - & - & - & 53\% \\
L56  & - & - & - & - & 87\% \\
L57  & - & - & - & - & 57\% \\
L58  & - & - & - & - & 50\% \\
L59  & - & - & - & - & 50\% \\
L60  & - & - & - & - & 57\% \\
L61  & - & - & - & - & 30\% \\
L62  & - & - & - & - & 43\% \\
L63  & - & - & - & - & 43\% \\
\hline
\end{tabular}
\end{minipage}
\end{table}

\begin{table}[ht]
\centering
\caption{Sensitivity study of re-selection layer (RL) on 10k-context-length Needle-in-the-Haystack test for Llama-3-8B-Instruct-Gradient-1048k with \Sys{}. The best accuracy for each token budget (K) is in bold. Layer 9, Layer 13, and Layer 14 are the best three re-selection layers for accuracy.}
\begin{tabular}{c|c|c|c|c|c|c}
\hline
\textbf{RL/K} & 16 & 32 & 64 & 128 & 256 & 512 \\
\hline
\Sys{}+L2  & 78\% & 84\% & 76\% & 94\% & 88\% & 98\% \\
\Sys{}+L3  & 0\% & 6\% & 10\% & 16\% & 28\% & 64\% \\
\Sys{}+L4  & 2\% & 10\% & 16\% & 28\% & 68\% & 84\% \\
\Sys{}+L5  & 10\% & 12\% & 32\% & 52\% & 72\% & 80\% \\
\Sys{}+L6  & 4\% & 6\% & 10\% & 14\% & 16\% & 24\% \\
\Sys{}+L7  & 2\% & 10\% & 10\% & 10\% & 14\% & 28\% \\
\Sys{}+L8  & 26\% & 64\% & 80\% & 90\% & 92\% & 96\% \\
\Sys{}+L9  & 52\% & 90\% & 96\% & \textbf{100\%} & 98\% & \textbf{100\%} \\
\Sys{}+L10 & 72\% & 76\% & 86\% & 94\% & 96\% & \textbf{100\%} \\
\Sys{}+L11 & 56\% & 74\% & 94\% & \textbf{100\%} & 98\% & 98\% \\
\Sys{}+L12 & 8\% & 14\% & 22\% & 44\% & 66\% & 94\% \\
\Sys{}+L13 & \textbf{92\%} & \textbf{92\%} & \textbf{96\%} & \textbf{100\%} & \textbf{100\%} & \textbf{100\%} \\
\Sys{}+L14 & 74\% & 68\% & 88\% & 98\% & \textbf{100\%} & \textbf{100\%} \\
\Sys{}+L15 & 74\% & 94\% & 92\% & 88\% & \textbf{100\%} & \textbf{100\%} \\
\Sys{}+L16 & 44\% & 50\% & 72\% & 66\% & 82\% & 94\% \\
\Sys{}+L17 & 42\% & 60\% & 74\% & 82\% & 96\% & 96\% \\
\Sys{}+L18 & 60\% & 72\% & 74\% & 74\% & 88\% & 98\% \\
\Sys{}+L19 & 58\% & 74\% & 82\% & 84\% & \textbf{98\%} & 96\% \\
\Sys{}+L20 & 64\% & 74\% & \textbf{96\%} & 78\% & 90\% & 98\% \\
\Sys{}+L21 & 10\% & 38\% & 60\% & 66\% & 90\% & 94\% \\
\Sys{}+L22 & 60\% & 70\% & 68\% & 72\% & 82\% & 98\% \\
\Sys{}+L23 & 58\% & 78\% & 70\% & 86\% & 88\% & 98\% \\
\Sys{}+L24 & 62\% & 58\% & 76\% & 70\% & 78\% & 92\% \\
\Sys{}+L25 & 66\% & 86\% & 84\% & 82\% & 92\% & \textbf{100\%} \\
\Sys{}+L26 & 54\% & 64\% & 66\% & 80\% & 90\% & 94\% \\
\Sys{}+L27 & 84\% & 80\% & 94\% & 96\% & 88\% & \textbf{100\%} \\
\Sys{}+L28 & 66\% & 66\% & 76\% & 84\% & \textbf{94\%} & 94\% \\
\Sys{}+L29 & 72\% & 80\% & 88\% & 80\% & 90\% & 96\% \\
\Sys{}+L30 & 80\% & \textbf{90\%} & 86\% & 88\% & 96\% & \textbf{100\%} \\
\Sys{}+L31 & 74\% & 90\% & 88\% & 84\% & 90\% & 96\% \\
\hline
\end{tabular}
\end{table}

\begin{table}[ht]
\centering
\caption{Sensitivity study of re-selection layer (RL) on 3k-context-length Needle-in-the-Haystack test for LongChat-7b-v1.5-32k with \Sys{}. The best accuracy for each token budget (K) is in bold. Layer 7 serves the best re-selection layer for accuracy.}
\begin{tabular}{c|c|c|c|c}
\hline
\textbf{RL/K} & 32 & 64 & 128 & 256 \\
\hline
\Sys{}+L2  & 2\%  & 2\%  & 6\%  & 54\%  \\
\Sys{}+L3  & 10\% & 52\% & 67\% & 78\%  \\
\Sys{}+L4  & 4\%  & 36\% & 65\% & 76\%  \\
\Sys{}+L5  & 17\% & 87\% & 94\% & 99\%  \\
\Sys{}+L6  & 70\% & 96\% & 99\% & 99\%  \\
\Sys{}+L7  & \textbf{80\%} & \textbf{98\%} & \textbf{100\%} & \textbf{100\%}  \\
\Sys{}+L8  & 58\% & 82\% & 96\% & 96\%  \\
\Sys{}+L9  & 7\%  & 31\% & 59\% & 71\%  \\
\Sys{}+L10 & 16\% & 59\% & 71\% & 78\%  \\
\Sys{}+L11 & 34\% & 61\% & 68\% & 77\%  \\
\Sys{}+L12 & 17\% & 32\% & 53\% & 77\%  \\
\Sys{}+L13 & 5\%  & 10\% & 28\% & 48\%  \\
\Sys{}+L14 & 24\% & 41\% & 57\% & 64\%  \\
\Sys{}+L15 & 37\% & 47\% & 62\% & 69\%  \\
\Sys{}+L16 & 16\% & 24\% & 28\% & 46\%  \\
\Sys{}+L17 & 4\%  & 4\%  & 10\% & 34\%  \\
\Sys{}+L18 & 2\%  & 3\%  & 8\%  & 15\%  \\
\Sys{}+L19 & 0\%  & 1\%  & 7\%  & 19\%  \\
\Sys{}+L20 & 0\%  & 3\%  & 6\%  & 20\%  \\
\Sys{}+L21 & 0\%  & 2\%  & 10\% & 19\%  \\
\Sys{}+L22 & 0\%  & 4\%  & 4\%  & 18\%  \\
\Sys{}+L23 & 0\%  & 2\%  & 5\%  & 13\%  \\
\Sys{}+L24 & 0\%  & 2\%  & 6\%  & 21\%  \\
\Sys{}+L25 & 0\%  & 2\%  & 7\%  & 16\%  \\
\Sys{}+L26 & 0\%  & 1\%  & 7\%  & 19\%  \\
\Sys{}+L27 & 0\%  & 2\%  & 4\%  & 21\%  \\
\Sys{}+L28 & 1\%  & 2\%  & 10\% & 17\%  \\
\Sys{}+L29 & 0\%  & 3\%  & 7\%  & 16\%  \\
\Sys{}+L30 & 1\%  & 1\%  & 9\%  & 15\%  \\
\Sys{}+L31 & 1\%  & 2\%  & 5\%  & 16\%  \\
\hline
\end{tabular}
\end{table}

\begin{table}[ht]
\centering
\caption{Sensitivity study of re-selection layer (RL) on 3k-context-length Needle-in-the-Haystack test for Yarn-Llama-2-7b-128k with \Sys{}. The best accuracy for each token budget (K) is in bold. Layer 7 serves the best re-selection layer for accuracy.}
\begin{tabular}{c|c|c|c|c|c}
\hline
\textbf{RL/K} & 32 & 64 & 128 & 256 & 512 \\
\hline
\Sys{}+L2  & 0\% & 0\% & 0\% & 3\% & 25\% \\
\Sys{}+L3  & 11\% & 26\% & 39\% & 65\% & 85\% \\
\Sys{}+L4  & 5\% & 17\% & 34\% & 60\% & 92\% \\
\Sys{}+L5  & 16\% & 42\% & 65\% & 87\% & 96\% \\
\Sys{}+L6  & 73\% & 83\% & 89\% & 95\% & \textbf{100\%} \\
\Sys{}+L7  & 73\% & \textbf{95\%} & \textbf{98\%} & \textbf{98\%} & \textbf{100\%} \\
\Sys{}+L8  & \textbf{87\%} & 92\% & 97\% & 94\% & 99\% \\
\Sys{}+L9  & 7\% & 21\% & 43\% & 60\% & 95\% \\
\Sys{}+L10  & 12\% & 31\% & 58\% & 69\% & 93\% \\
\Sys{}+L11  & 20\% & 21\% & 46\% & 68\% & 97\% \\
\Sys{}+L12  & 2\% & 15\% & 28\% & 51\% & 92\% \\
\Sys{}+L13  & 4\% & 5\% & 20\% & 34\% & 88\% \\
\Sys{}+L14  & 16\% & 20\% & 49\% & 53\% & 91\% \\
\Sys{}+L15  & 2\% & 25\% & 44\% & 56\% & 90\% \\
\Sys{}+L16  & 10\% & 13\% & 21\% & 43\% & 86\% \\
\Sys{}+L17  & 3\% & 4\% & 9\% & 16\% & 85\% \\
\Sys{}+L18  & 0\% & 1\% & 2\% & 15\% & 84\% \\
\Sys{}+L19  & 0\% & 2\% & 3\% & 7\% & 80\% \\
\Sys{}+L20  & 0\% & 2\% & 0\% & 7\% & 79\% \\
\Sys{}+L21  & 0\% & 0\% & 1\% & 5\% & 77\% \\
\Sys{}+L22  & 0\% & 0\% & 0\% & 8\% & 76\% \\
\Sys{}+L23  & 0\% & 0\% & 1\% & 7\% & 74\% \\
\Sys{}+L24  & 0\% & 2\% & 0\% & 9\% & 73\% \\
\Sys{}+L25  & 0\% & 0\% & 2\% & 10\% & 71\% \\
\Sys{}+L26  & 0\% & 1\% & 1\% & 5\% & 70\% \\
\Sys{}+L27  & 0\% & 1\% & 3\% & 8\% & 68\% \\
\Sys{}+L28  & 0\% & 1\% & 0\% & 9\% & 67\% \\
\Sys{}+L29  & 0\% & 0\% & 2\% & 8\% & 65\% \\
\Sys{}+L30  & 0\% & 1\% & 1\% & 8\% & 64\% \\
\Sys{}+L31  & 0\% & 0\% & 2\% & 10\% & 62\% \\
\hline
\end{tabular}
\end{table}

\end{document}